\pdfoutput=1

\documentclass[11pt]{article}

\usepackage[final]{EMNLP2022}

\usepackage{times}
\usepackage{amsmath}
\usepackage{latexsym}
\usepackage{booktabs}       
\usepackage{multirow}
\usepackage{graphicx}
\usepackage{soul}
\usepackage{xcolor}
\usepackage{rotating}
\usepackage[framemethod=tikz]{mdframed}  
\usepackage{enumitem}

\usepackage[T1]{fontenc}

\usepackage[utf8]{inputenc}
\usepackage{microtype}
\usepackage{hyperref}

\newcommand*\samethanks[1][\value{footnote}]{\footnotemark[#1]}

\title{What Language Model to Train if You Have One Million GPU Hours?}

\author{\textbf{\underline{The BigScience Architecture \& Scaling Group}} \vspace{0.4cm}\\ \small
\textbf{Teven Le Scao}$^{1}$\thanks{~~Equal contribution.} \hspace{0.3cm}
\textbf{Thomas Wang}$^{1}$\samethanks \hspace{0.3cm}
\textbf{Daniel Hesslow}$^{2}$\samethanks \hspace{0.3cm}  \textbf{Lucile Saulnier}$^{1}$\samethanks \hspace{0.3cm}
\textbf{Stas Bekman}$^{1}$\samethanks \\
\small
\textbf{M Saiful Bari}$^3$ \hspace{0.3cm} \textbf{Stella Biderman}$^{4,5}$ \hspace{0.3cm} \textbf{Hady Elsahar}$^6$ \hspace{0.3cm} 
\textbf{Niklas Muennighoff}$^1$ \hspace{0.3cm} 
\textbf{Jason Phang}$^5$ \hspace{0.3cm} \textbf{Ofir Press}$^8$ \\
\small
\textbf{Colin Raffel}$^1$ \hspace{0.3cm}
\textbf{Victor Sanh}$^1$ \hspace{0.3cm}
 \textbf{Sheng Shen}$^9$ \hspace{0.3cm} \textbf{Lintang Sutawika}$^{10}$ \hspace{0.3cm} \textbf{Jaesung Tae}$^1$ \hspace{0.3cm} \textbf{Zheng Xin Yong}$^{11}$ \\
 \small
 \textbf{Julien Launay}$^{2, 12}$\thanks{~~Equal supervision.} \hspace{0.3cm}
 \textbf{Iz Beltagy}$^{13}$\samethanks\vspace{0.1cm} \\
 \small
 $^1$ Hugging Face \hspace{0.2cm} $^2$ LightOn \hspace{0.2cm} $^3$ NTU, Singapore \hspace{0.2cm} $^4$ Booz Allen \hspace{0.2cm} $^5$ EleutherAI \hspace{0.2cm} $^6$ Naver Labs Europe \hspace{0.2cm} $^7$ New York University\\
 \small$^8$ University of Washington \hspace{0.2cm} $^9$ Berkeley University \hspace{0.2cm} $^{10}$ Big Science \hspace{0.2cm} $^{11}$ Brown University \hspace{0.2cm} $^{12}$ LPENS \hspace{0.2cm} $^{13}$ Allen Institute for AI
}

\begin{document}
\onecolumn
\maketitle
\begin{abstract}
The crystallization of modeling methods around the Transformer architecture has been a boon for practitioners. 
Simple, well-motivated architectural variations can transfer across tasks and scale, increasing the impact of modeling research. 
However, with the emergence of state-of-the-art 100B+ parameters models, large language models are increasingly expensive to accurately design and train. 
Notably, it can be difficult to evaluate how modeling decisions may impact emergent capabilities, given that these capabilities arise mainly from sheer scale alone.
In the process of building BLOOM--the Big Science Large Open-science Open-access Multilingual language model--our goal is to identify an architecture and training setup that makes the best use of our 1,000,000 A100-GPU-hours budget.
Specifically, we perform an ablation study at the billion-parameter scale comparing different modeling practices and their impact on zero-shot generalization.
In addition, we study the impact of various popular pre-training corpora on zero-shot generalization. 
We also study the performance of a multilingual model and how it compares to the English-only one. 
Finally, we consider the scaling behaviour of Transformers to choose the target model size, shape, and training setup. All our models and code are open-sourced at \url{https://huggingface.co/bigscience}.
\end{abstract}
\section{Introduction}
\begin{figure}
    \centering
    \includegraphics[width=\columnwidth]{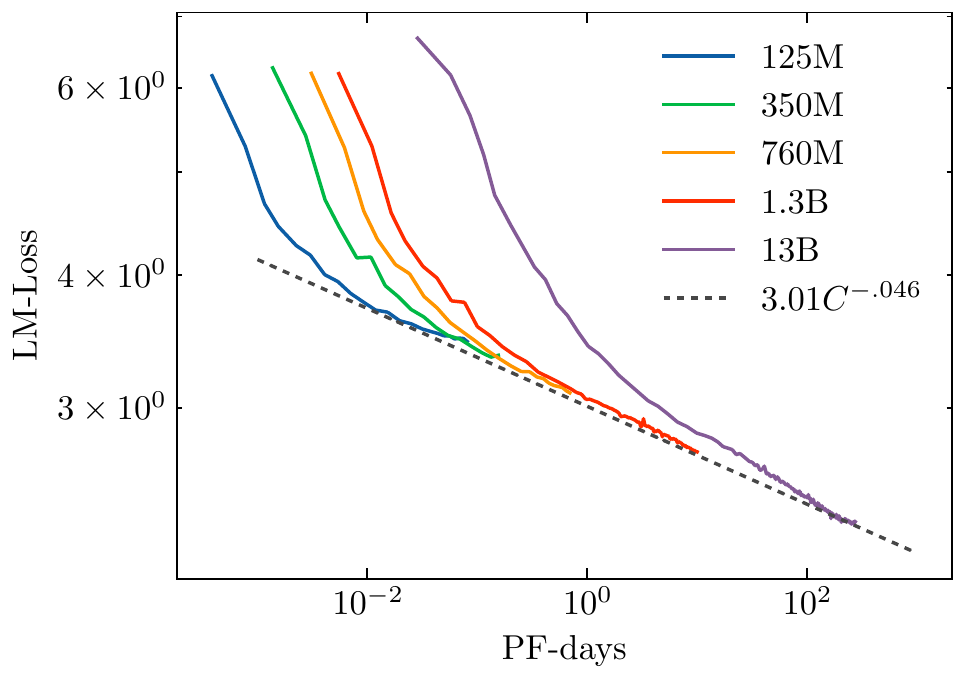}
    \caption{\textbf{Smooth scaling of language modeling loss as compute budget and model size increase}. We observe a power-law coefficient $\alpha_C \sim 0.046$, in-line with \citet{kaplan2020scaling}. We use this fit to estimate the optimal size and number of tokens to train on for the final model given the available budget.}
    \label{fig:scaling}
\end{figure}

Recent years have seen the advent of large language models 
characterized by emergent capabilities (e.g., zero-shot generalization) arising from sheer scale alone~\cite{radford2019language,brown2020gpt3}.
Scaling LLMs results in a predictable increase in performance: simple scaling laws connect the number of parameters, pretraining dataset size, and compute budget~\cite{kaplan2020scaling,ganguli2022predictability,hoffmann2022training}, providing a clear path towards more capable models. This paradigm shift has been fueled by the wide~adoption of the Transformer~\cite{vaswani2017attention}, providing a scalable basis for practitioners to build upon. 

In this paper, we design an architecture and training setup for a multilingual 100B+ parameters model (BLOOM, \citet{bigscience_workshop_2022}), seeking to best use a fixed 1,000,000 A100-hours budget. Because of the costs involved with training large language models, we cannot exhaustively explore the landscape of possible models. Instead, we position ourselves as practitioners exploring "off-the-shelf" solutions. We thus test promising additions to the Transformer to attempt to reproduce their findings in a controlled, large-scale setting.

Although our main goal was to prepare the architecture and training setup of BLOOM, our findings are also valuable for practitioners building models in the 1-10B range, as they equally improve the performance of such smaller models. At variance with major works on large language models, we also make a significant effort towards reproducibility and openness: all of our pretrained models, code, and notes from our weekly meetings are made available. See Appendix \ref{sec:artefacts} for the relevant links.

\paragraph{Contributions.} We first study the impact of pretraining corpora, positional embeddings, activation functions, and embedding norm on zero-shot generalization. We base our study on the popular GPT-2 architecture \cite{radford2019language}, with experiments at the 1.3B parameters scale. We then consider the impact of massive multilinguality, showing language-specific scaling laws in a multilingual setting for the first time. Finally, we describe our approach to drafting an architecture for the final 176B parameters BLOOM model.

\section{Methods}

\begin{table*}[t]
\begin{center}
\begin{tabular}{@{}lccccc@{}}
\toprule
\multicolumn{1}{c}{\textbf{Model}} & \textbf{Parameters}            & \multicolumn{4}{c}{\textbf{Pretraining tokens}}                                                                       \\ \midrule
                                               & \multicolumn{1}{l}{}  & \multicolumn{1}{l}{Dataset} & \multicolumn{1}{l}{112B} & \multicolumn{1}{l}{250B} & \multicolumn{1}{l}{300B} \\ \midrule
\textbf{OpenAI} --- Curie          & 6.7B            & \multicolumn{1}{l}{}        &                          &                          & \underline{49.28}           \\
\textbf{OpenAI} --- Babbage          & 1.3B            & \multicolumn{1}{l}{}        &                          &                          & \textbf{45.30}        \\
\textbf{EleutherAI} --- GPT-Neo              & 1.3B                  & The Pile                    &                          &                          & 42.94                    \\ \midrule
\multirow{1}{*}{\textbf{Ours}}                             & 13B                   & OSCAR v1                      &                          &                          & 47.09                    \\ \midrule
    \multirow{3}{*}{\textbf{Ours}}                                                        & 1.3B & The Pile                    & \textbf{42.79}                    & 43.12                    & 43.46                    \\
                                                           & 1.3B & C4                          & 42.77                    &                          &                          \\
                                                           & 1.3B & OSCAR v1                       & 41.72           &                          &                          \\ \bottomrule
\end{tabular}
\end{center}
\caption{\textbf{Pretraining datasets with diverse cross-domain high-quality data improves zero-shot generalization.} Average accuracy on EAI harness (higher is better) using different pretraining corpora and comparison with baseline models. \textbf{Bold is best 1.3B model for amount of tokens seen}, \underline{underline is best overall}.}
\label{tab:validation}
\end{table*}

We first justify our choice to base our model on the popular recipe of combining a decoder-only model with an autoregressive language modeling objective, and introduce our experimental setup. We then discuss our evaluation benchmarks, and motivate our choice of zero-shot generalization as our key metric. Finally, we introduce the baselines we compare to throughout the paper.

\subsection{Architecture and Pretraining Objective}
\label{sec:t5x}

In this paper, we base all models on a decoder-only Transformer pretrained with an autoregressive language modeling objective. This is a popular choice for large language models \cite{brown2020gpt3, rae2021scaling, thoppilan2022lamda}, possibly because it lends itself to zero-shot application to many downstream tasks \cite{radford2019language}. Alternatives include encoder-decoder models trained with a span-corruption objective (e.g., T5~\citet{raffel2019t5}), as well as non-causal decoders models with visibility over a prefix (so-called Prefix LMs, \citet{liu2018generating, dong2019unified}).   

Our decision is motivated by the findings of~\citet{wang2022language}, which showed that decoder-only models combined with an autoregressive language modeling objective provide the best zero-shot generalization abilities immediately after pretraining. Although multitask finetuning~\cite{Sanh2021MultitaskPT,wei2021finetuned} will instead favor an encoder-decoder with span corruption for best zero-shot generalization, \citet{wang2022language} found a compromise between these two practices. Following autoregressive pretraining, decoder-only models can be efficiently adapted into non-causal decoders, simply by extending pretraining with span corruption. This adaptation produces a second model, which can provide excellent zero-shot generalization after multitask finetuning. Accordingly, we follow their recommendation, and train an autoregressive decoder-only model first which we will later consider adapting and finetuning. 

\subsection{Experimental Setup} 
We follow the architectures GPT-2~\citep{radford2019language} and the hyperparameters of GPT-3 \citep{brown2020gpt3}. For learning rate, we use a maximum value of $2 \times 10^{-4}$, with a linear warm-up over 375M tokens, followed by cosine decay to a minimum value of $1 \times 10^{-5}$. We use a 1M tokens batch size, with linear ramp-up over the first 4B tokens, and a sequence length of 2,048. We use the Adam optimizer \cite{kingma2014adam}, with $\beta_1=0.9$, $\beta_2=0.999$, $\epsilon=1 \times 10^{-8}$, weight decay 0.1, and gradient clipping to 1.0. We also tie the word embedding and softmax matrix~\citep{tying}. Unless noted otherwise, we conduct our experiments with 1.3B parameters models, pretraining on 112B tokens. 

We picked this size and dataset size as a compromise between compute cost and the likelihood that our conclusions would transfer to the target 100B+ model. Notably, we needed to be able to reliably measure zero-shot generalization above random chance. We note that training for 112B tokens 1.3B parameters models bring them significantly above the optimality threshold of~\citet{kaplan2020scaling}, and of~\citet{hoffmann2022training}.

The main architectural difference with GPT-3 is that all our layers use full attention, while GPT-3 uses alternating sparse attention layers~\citep{sparse}. 
The main value of sparse attention layers is to save compute with long sequence lengths. However, at the 100B+ scale, sparse attention layers
provide negligible compute savings, as the vast majority of the compute is spent on the large feed-forward layers.
\citet{kaplan2020scaling} estimated the amount of compute per token to be:
\begin{equation*}
C_\text{forward} = 2 \times (12 n_\text{layer} d^2 + n_\text{layer} n_\text{ctx} d),
\end{equation*}
where $C_\text{forward}$ is the cost for the forward pass, $n_\text{layer}$ is the number of layers, $d$ is the hidden dimension, and $n_\text{ctx}$ is the sequence length. This means if $12 d >> n_\text{ctx}$, the 
second $n_\text{layer} n_\text{ctx} d$ term is negligible, which is the case for our final model 
where $d > 10,000$ and $n_\text{ctx} = 2048$. 

\paragraph{What is a FLOP exactly?} We report throughput per GPU in FLOPS and total budgets in PF-days (i.e. one PFLOPS sustained for a day). It is important to highlight that FLOPS are never directly measured, but always estimated, with widely different practices across papers. We refer to \emph{model} FLOP the estimates based on the $C=6ND$ formula from \citet{kaplan2020scaling}, where $C$ is the total compute, $N$ the model size, and $D$ the number of tokens processed. These are the FLOP actually used to train the model, and which are used for scaling laws. We refer to \emph{hardware} FLOP the estimates reported by our codebase, using the formula from \citet{narayanan2021efficient}. This notably includes gradient checkpointing, which trades additionnal computations for reduced memory needs, and a more thorough accounting of operations.

\subsection{Evaluation Benchmarks} 
We measure upstream performance using the language modeling loss on an held out sample of the pretraining dataset. However, it is not always possible to compare losses across objectives and tokenizers. Moreover, as upstream performance is not always aligned with task performance \cite{Tay2021ScaleEI}, we must also measure downstream performance explicitly. We could use zero/few-shot generalization, with or without specific finetuning. 

Specifically, we choose to measure zero-shot generalization on a diverse set of tasks. Few-shot and zero-shot results are strongly correlated: we found a Pearson correlation coefficient of 0.93 between zero-shot and few-shot performance across model sizes in \citet{brown2020gpt3}. We do not rely on finetuning as it is not how the main final model is likely to be used, given its size and the challenges associated with finetuning at the 100B+ scale. 

We use the popular EleutherAI Language Model Evaluation Harness (EAI harness, \citet{eval-harness}), evaluating models across 27 diverse tasks that are similar to those used in~\citet{brown2020gpt3} (see Appendix \ref{sec:sup_eval} for a list of tasks). Overall, the random baseline on our benchmark sits at 33.3\%. 

\subsection{Baselines} 
We use GPT-Neo~\cite{gpt-neo}, a~1.3B decoder-only autoregressive language model trained on the Pile~\cite{gao2020pile}, and GPT-3~\cite{brown2020gpt3}, accessed via the OpenAI API. We evaluate two models, Babbage and Curie\footnote{These models are now referred to as \texttt{text-babbage-001} and \texttt{text-curie-001}.}. Based on \citet{gaosize} and our own analysis, we assume  
Babbage is 1.3B while Curie is 6.7B based on how close our computed results are to those reported in the original paper. However, as details of the OpenAI API are kept secret, there is no way to make sure that the models are actually the ones described in~\citet{brown2020gpt3} -- the number of pretraining tokens reported in Table \ref{tab:validation} is thus to be taken cautiously.
\section{Impact of Pretraining Data}

We first study the impact of pretraining data on zero-shot generalization. More diverse pretraining data, ideally curated from a cross-domain collection of high-quality datasets, has been suggested to help with downstream task performance and zero-shot generalization \cite{rossettnlg, gao2020pile}. 

\subsection{Corpora} 
We evaluate three possible corpora, all commonly used to train large language models:
\begin{itemize}
    \item \textbf{OSCAR v1}~\citep{ortiz2019oscar}\footnote{The recent release of OSCAR v2 is a better dataset, but it wasn't available when we started this project.}, a multilingual, filtered version of Common Crawl;
    \item \textbf{C4} \citep{raffel2019t5}, specifically its replication by AllenAI, a processed and filtered version of Common Crawl;
    \item \textbf{The Pile} \citep{gao2020pile}, a diverse pretraining corpus that contains webscrapes from Common Crawl in addition to high-quality data from cross-domain sources such as academic texts and source code.
\end{itemize}

For each pretraining corpus, we train a 1.3B parameter model for 112B tokens. For the Pile specifically, motivated by good early results at 112B tokens, we train up to 300B tokens, to compare with GPT-3 models and validate against GPT-Neo. 

\subsection{Results}

Evaluation results are outlined in Table \ref{tab:validation}. We find that training on the Pile produces models that are better at zero-shot generalization, with C4 a close second, and OSCAR significantly behind. 

Importantly, this finding transfers to larger scales: as part of engineering test runs, a 13B model was trained on OSCAR for 300B tokens. 
We found this 13B model to underperform the 6.7B model from OpenAI API 
which we attribute to the low quality of the English data in OSCAR. 

We also note that our model trained on The Pile outperforms the 1.3B GPT-Neo trained on the same dataset. Finally, our 1.3B model still underperforms the 1.3B model from the OpenAI API by 1.6\%. It seems most likely that the difference is that of data, but we cannot investigate this further as the GPT-3 training dataset is neither publicly available nor reproducible.

\begin{mdframed}
\textbf{Finding 1.} Diverse cross-domain pretraining data combining web crawls with curated high-quality sources improves zero-shot generalization over pretraining datasets constructed from Common Crawl only.
\end{mdframed}

\section{Architecture Ablations}
We now consider ablation studies to better identify the best positional embedding, activation function, and embedding normalization placement. 

\subsection{Positional Embeddings}

\begin{table}[b]
\begin{center}
\begin{tabular}{@{}cc@{}}
\toprule
\textbf{Positional Embedding} & \textbf{Average EAI Results}\\
\midrule
None & 41.23\\
Learned & 41.71\\
Rotary & 41.46\\
ALiBi & \textbf{43.70} \\ 
\bottomrule
\end{tabular}
\end{center}
\caption{\textbf{ALiBi significantly outperforms other embeddings for zero-shot generalization.} All models are trained on the OSCAR dataset for 112 billion tokens.}
\label{tab:positional}
\end{table}

\paragraph{Background} Originally, both static sinusoidal position embeddings and learned position embeddings were proposed to capture positionnal information; the latter are popular in large language models \cite{brown2020gpt3}. 
\citet{su2021roformer} proposed rotary embeddings, where the query and key representations inside the self-attention mechanism are modified such that the attention captures relative distances between them. Recently, \citet{press2021alibi} introduced a method which does not use embeddings, instead directly attenuating the attention scores based on how far away the keys/queries are. 

\begin{figure}[h]
    \centering
    \includegraphics[width=\columnwidth]{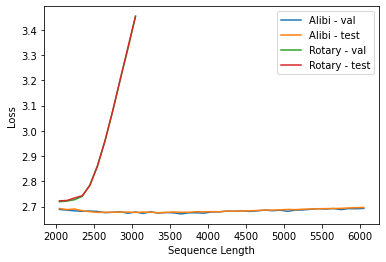}
    \caption{\textbf{ALiBi embeddings can effectively extrapolate past the sequence length on which the model was trained, while rotary embeddings can not.} This is in line with the findings of \citet{press2021alibi}.}
    \label{fig:extrapolation}
\end{figure}

\paragraph{Results} We compare learned, rotary, and ALiBi position embeddings, and include a baseline without position embeddings. Our results are presented in Table~\ref{tab:positional}. Although learned positional embeddings outperform rotary embeddings, ALiBi yields significantly better results than all alternatives. We also confirm the findings of~\citet{biderman2021nopos}: a baseline with no positional information exhibits competitive performance. While bidirectional models require positional embeddings to determine the location of tokens, we find autoregressive models can simply leverage the causal attention mask. We also confirm the ability of ALiBi to extrapolate to longer sequences than trained on in Figure \ref{fig:extrapolation}. Note that results in Table~\ref{tab:positional} do not use any extrapolation: ALiBi embeddings are a better choice even without taking into account their ability to extrapolate. 

\begin{table}[b]
\begin{center}
\begin{tabular}{@{}cc@{}}
\toprule
\textbf{Activation function} & \textbf{Average EAI Results}\\
\midrule
GELU & 42.79\\
SwiGLU & \textbf{42.95}\\
\bottomrule
\end{tabular}
\end{center}
\caption{\textbf{SwiGLU slightly outperforms GELU for zero-shot generalization.} Models trained on The Pile for 112 billion tokens.}
\label{tab:activation}
\end{table}

\begin{mdframed}
\textbf{Finding 2.} ALiBi positional embeddings significantly outperforms other embeddings for zero-shot generalization.
\end{mdframed}

\subsection{Activation Functions}

\paragraph{Background.} Large language models by and large still mostly use the GELU activation \cite{hendrycks2016gaussian}. We evaluate a recently proposed alternative, SwiGLU \cite{shazeer2020swiglu}, which combines both Gated Linear Units \cite{dauphin2016glu} with the Swish activation function \cite{ramachandran2017searching}. 

SwiGLU uses $50\% $ extra parameters in the feed-forward layers. As suggested in \citet{shazeer2020swiglu}, we compensate for this by reducing the hidden size of the feed-forward layer. 

\paragraph{Results.} We present our results in Table~\ref{tab:activation}. SwiGLU produces slightly better results than GELU. For our final model, we adopted GELU, as we initially observed a lower throughput for SwiGLU. However, further benchmarking identified that this overhead was primarily associated with the change in the hidden size of the feedforward network. Indeed, this new size, 5,456, is divisible by neither the warp size of the GPU~\citep{Lashgar2013WarpSI} nor the number of streaming multiprocessors, resulting in both tile and wave quantization. We accordingly recommend using SwiGLU for future models.

\subsection{Embedding Norm}

\citet{bitsandbytes} suggests that greater stability of training can be achieved by including an extra layer normalization \cite{layernorm} after the embedding layer. We evaluate the performance impact of such a modification in Table~\ref{tab:emb_norm}. We note that this incurs a significant reduction in the performance of the model. However, models above 100 billion parameters are notoriously unstable and require considerable engineering efforts in order to be kept stable. If this addition provides increased stability when training, it may be valuable.

\begin{table}[b]
\begin{center}
\begin{tabular}{@{}cc@{}}
\toprule
\textbf{Embedding Norm} & \textbf{Average EAI Results}\\
\midrule
No & \textbf{43.46}\\
Yes &  42.24\\
\bottomrule
\end{tabular}
\end{center}
\caption{\textbf{Layer normalization after the embedding layer diminishes performance significantly.} Models trained on The Pile for 300 billion tokens.}
\label{tab:emb_norm}
\end{table}

\begin{mdframed}
\textbf{Finding 3.} Adding layer normalization after the embedding layer incurs a significant penalty on zero-shot generalization. 
\end{mdframed}

\begin{table*}[t!]
\begin{center}
\begin{tabular}{@{}rc|cccccccc|c@{}}
\toprule
\textbf{Model} &  \textbf{Size} & EN & ZH & ES & FR & VI & AR & HI & UR & \textbf{Average} \\
\midrule
XGLM~(\citeauthor{XGLM}) & 7.5B & 54.5 & 45 & 38.2 & 50.7 & 47.5 & 47.5 & 43.4 & 42.7 & 46.19 \\
XGLM (reprod.)  & 7.5B & 53.85 & 45.21 & 41.7 & 49.82 & 47.35 & 46.37 & 43.19 & 42.3 & 46.22 \\
XGLM  & 1.7B & 49.68 & 44.63 & 37.39 & 47.94 & 42.75 & 45.65 & 44.35 & 43.19 & 44.45 \\
Ours  & 1.3B & 49.9 & 44.53 & 36.77 & 46.51 & 45.75 & 43.41 & 45.95 & 42.91 & 44.47\\
\bottomrule
\end{tabular}
\end{center}
\caption{\textbf{Our multilingual 1.3B model achieves accuracy on zero-shot XNLI in line with XGLM~\citet{XGLM}.} First row is the reported XGLM results, and the second is our reproduction of their results to validate our multilingual evaluation setup. Last two rows show that our multilingual model matches the XGLM results. } 
\label{tab:mutlilingual_xnli}
\end{table*}

\section{Multilinguality}
\begin{table}[b]
\begin{center}
\begin{tabular}{@{}cc@{}}
\toprule
\textbf{Pretraining} & \textbf{Average EAI Results}\\
\midrule
English-only  & \textbf{41.72}\\
Multilingual  & 38.55\\
\bottomrule
\end{tabular}
\end{center}
\caption{\textbf{Multilingual pretraining very significantly diminishes English zero-shot generalization.} Both models trained on OSCAR for 112B tokens.}
\label{tab:mutlilingual}
\end{table}

The majority of 100B+ language models have been trained in English, with notable exceptions in Chinese~\citep{zeng2021pangu, wu2021yuan} and Korean~\cite{Kim2021WhatCC} models. Smaller massively multilingual models have seen wider adoption \cite{mT5}, but these models are not suitable for zero-shot. Recent results on large GPT-like multilingual models show that English-only performance is usually disappointing \cite{XGLM}. 

\paragraph{Training data.}
We train a multilingual model to evaluate the effectiveness and potential impacts of this practice. 
We use the OSCAR dataset~\citep{ortiz2019oscar}, but here we include multiple languages, not only English as in the earlier experiments. 
The languages we include are Arabic, Basque, Bengali, Chinese, Catalan, English, French, Hindi, Indonesian, Portuguese, Spanish, Urdu, and Vietnamese.
We sample each language with a different probability that downsamples the most frequent languages 
and upsamples the least frequent ones, so that all languages 
are represented. We estimate the sampling probabilities similar to~\citet{Xue2021mT5AM}.

\paragraph{English-only evaluation.}
We first evaluate our multilingual model on the same set of English benchmarks we have used previously, in Table~\ref{tab:mutlilingual}. Multilinguality significantly lowers accuracy on the English benchmark, which is in line with the results from~\citet{XGLM}. 

\paragraph{Multilingual evaluation.}
Zero-shot multilingual evaluation is more challenging to setup because it requires writing new prompts for each new language. Therefore, instead of manually writing prompts for each language, we follow the strategy proposed by~\citet{XGLM}, using English prompts for non-English examples--this can be viewed as cross-lingual zero-shot generalization. They validated this strategy by demonstrating its ability to achieve zero-shot performance on par with (and sometimes even better than) human-written language-specific prompts. This strategy also demonstrates cross-lingual abilities.

\definecolor{shadecolor}{rgb}{0.93,0.93,0.93}

We evaluate on XNLI~\citep{conneau2018xnli}, a multilingual NLI dataset  that covers 8 of the languages we use for training. 
Our evaluation is different from the zero-shot evaluation of the XTREME benchmark~\cite{Hu2020XTREMEAM}. XTREME first finetunes the model on the English training data of each downstream task, then evaluates it on the non-English dataset, attempting cross-lingual generalization. 
Our evaluation avoids any finetuning, and instead relies entirely on zero-shot generalization.

\paragraph{Results.}
Table~\ref{tab:mutlilingual_xnli} shows the XNLI results of our multilingual model and how it compares to XGLM~\cite{XGLM}.
We were able to reproduce the results of XGLM-7.5B which validates our evaluation setup. Furthermore, the table shows that the performance of our 1.3B 
is in line with the XNLI 1.7B model, validating that our multilingual setup achieves competitive results. It is worth noting that our 1.3B model is trained on only 112B tokens from 13 languages while 
XGLM is trained on 500B tokens from 30 languages. As far as we are aware, this is the first independent replication of the main results of~\citet{XGLM}.

\begin{figure*}[t]
    \centering
    \centerline{\includegraphics[width=1.1\textwidth]{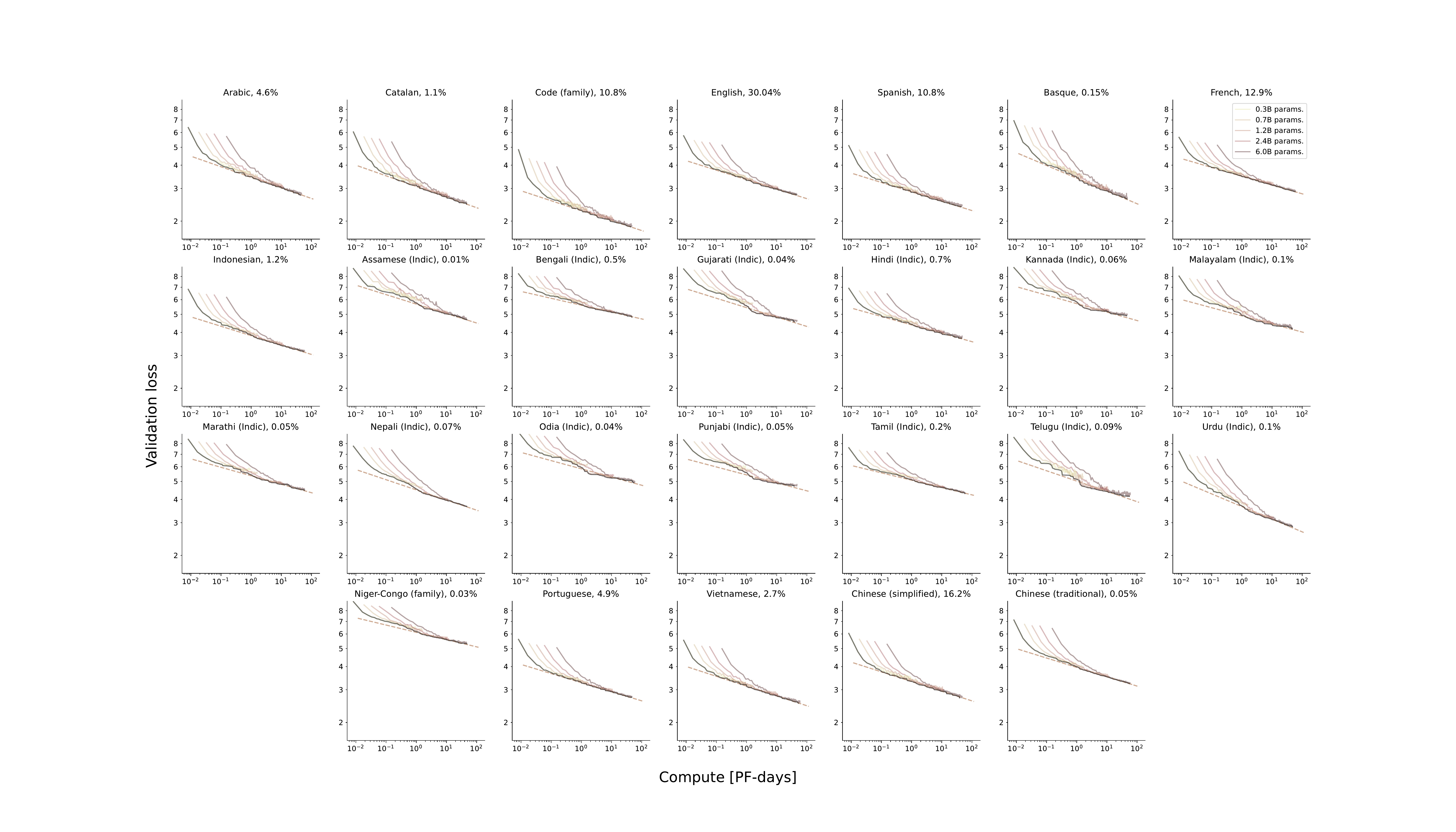}}
    \caption{\textbf{Scaling laws across languages for the smaller BLOOM models}. Black line is Pareto frontier of optimality (best loss at a given compute), dashed line is best fit. Fit coefficients are detailed in Appendix \ref{sec:multilingualscalinglaws}. All sufficiently represented languages exhibit similar scaling behaviour, with mostly differences in loss offsets.}
    \label{fig:multilingualscaling}
\end{figure*}

\paragraph{Language-specific scaling laws.} To explore how scale influences multilinguality, we train a wider range of models (i.e. 0.3-6B parameters) on a larger corpus of more than 300B tokens of text drawn from a variety of languages \cite{roots}. In Figure \ref{fig:multilingualscaling}, we show scaling laws for Arabic, Catalan, Code, English, Spanish, Basque, French, Indonesian, Assamese, Bengali, Gujarati, Hindi, Kannada, Malayalam, Marathi, Nepali, Odia, Punjabi, Tamil, Telugu, Urdu, aggregated Niger-Congo languages, Portuguese, Vietnamese, Simplified and Traditional Chinese. 

Smaller models struggle more with under-represented languages such as those in the Indic and Niger-Congo family. For example, the loss of the sub-1 billion models goes up at the end of training for Malayalam, Odia, and Telugu. As data is not repeated, it is unlikely that this effect is due to overfitting; we interpret this as insufficient capacity in the model to handle many language representations, with data in the dominant language sets causing catastrophic forgetting of less represented languages. In contrast, the largest model sees its loss decrease smoothly for every language: larger models handle multilinguality more easily. Overall, scaling laws coefficients are consistent across well-represented languages, only differing in offsets.

\section{Scaling to 176B parameters}
We now detail how our previous findings influence our architecture and scaling decisions for the final 176B BLOOM model. 

\paragraph{Compute allocation.} We have been allocated 18 weeks of dedicated use of partition with 52 nodes of 8x 80GB A100 GPUs on the Jean Zay supercomputer. We set four nodes aside as spare, so that our compute budget amounts to 1,161,216 A100-hours in total. Assuming a throughput of 100 model TFLOPS, approximately corresponding to state-of-the-art hardware FLOPS of 150~\cite{narayanan2021efficient}, we have a compute budget of 4,838 PF-days for the model training. We round this down to 4,500 PF-days, this $\sim10$\% safety margin accounting for potential downtime and inefficiencies (e.g., batch size ramp-up) during training. To put this number in perspective, this is $\sim23$\% more than the training budget of GPT-3. Given this compute budget, our English-only scaling laws in \ref{fig:scaling} predict an optimal allocation for training a 392B parameter model for 165B tokens. We will use these as an upper bound in size: the largest model we can afford is 392B parameters, and the minimum number of tokens to train on is 165B tokens.

\begin{table*}[t]
\begin{small}
\begin{center}
\begin{tabular}{@{}lccccccc@{}}
\toprule
\multicolumn{1}{c}{\textbf{Model}} & \textbf{Size} & \textbf{Pretraining} & \textbf{Budget} & \textbf{Layers} & \textbf{Hidden dim.} & \multicolumn{2}{c}{\textbf{Attention heads}} \\ 
               & {[Bparams.]} & {[Btokens]}         & {[PF-days]}   &                 &                      & num.                  & dim.                 \\ \midrule
LaMDA \cite{thoppilan2022lamda}          & 137           & 432                  & 4,106            & 64              & 8,192                & 128                   & 64                   \\
GPT-3 \cite{brown2020gpt3}         & 175           & 300                  & 3,646            & 96              & 12,288               & 96                    & 128                  \\
J1-Jumbo \cite{J1WhitePaper}       & 178           & 300                  & 3,708            & 76              & 13,824               & 96                    & 144                  \\
PanGu-$\alpha$ \cite{zeng2021pangu} & 207           & 42                   & 604              & 64              & 16,384               & 128                   & 128                  \\
Yuan \cite{wu2021yuan}           & 245           & 180                  & 3,063            & 76              & 16,384               &  & \\
Gopher \cite{rae2021scaling}         & 280           & 300                  & 4,313            & 80              & 16,384               & 128                   & 128                  \\
MT-530B \cite{smith2022using}       & 530           & 270                  & 9,938            & 105             & 20,480               & 128                   & 160                  \\ \bottomrule
\end{tabular}
\end{center}
\end{small}
\caption{\textbf{State-of-the-art 100B+ models with publicly available details.} Compute budget is expressed in model PF-days required for training the models, from the $C= 6ND$ approximation of \citet{kaplan2020scaling}. Number of tokens for LaMDA is inferred from reported compute budget and size. Yuan did not report attention head details.}
\label{tab:sota_models}
\end{table*}

\begin{table*}[t]
\begin{small}
\begin{center}
\begin{tabular}{ccccccccc}
\toprule
\multicolumn{1}{c}{\textbf{Model}} & \textbf{Size} & \textbf{Layers} & \textbf{Hidden dim.}    & \multicolumn{2}{c}{\textbf{Attention heads}} & \multicolumn{1}{c}{\textbf{Memory}} & \multicolumn{2}{c}{\textbf{Performance}}                       \\
\multicolumn{1}{c}{}               & [params.]     &                 &                         & num.                  & dim.                 & \multicolumn{1}{c}{[GB]}            & \multicolumn{1}{c}{[sec/iter.]} & \multicolumn{1}{c}{[TFLOPs]} \\ \midrule
(1)                                & 178           & 82              & \multirow{2}{*}{13,312} & 64                    & 208                  & 63                                  & 104                             & 152                          \\
(2)                                & 178           & 82              &                         & 128                   & 104                  & 60                                  & 109                             & 146                          \\
\textbf{(3)}                                & \textbf{176}           & \textbf{70}              & \textbf{14,336}                  & \textbf{112}                   & \textbf{128}                  & \textbf{59}                                  & \textbf{105}                             & \textbf{150}                          \\ \bottomrule
\end{tabular}
\end{center}
\end{small}
\caption{\textbf{We choose configuration (3) as the final configuration for our 176B model.} (1) was rejected because of high attention heads dimension, and (3) was favored over (2) because of higher throughput. Appendix \ref{sec:arch_details} details all 20 final configurations benchmarked, only the best three are displayed here.}\label{tab:final_configs}
\end{table*}

\paragraph{Model shape.} 
\citet{kaplan2020scaling} studied the dependence of the loss with model shape, and found only a limited impact within a wide range of feed-forward ratios $d_{ff} / d_{model}$, aspect ratios $d_{model}/n_{layer}$, and attention head dimensions. 

\citet{levine2020limits} proposed a theoretically motivated and empirically backed law describing the optimal compromise between width and depth. They predict that 100B+ parameters models such as GPT-3 are too deep, while models in the 10B or smaller range are usually too shallow. For a GPT-3-sized model with 175B parameters, they predict an ideal depth of 80 layers. 

\subsection{Final Model Architecture}

We set three main guidelines for our final model:
\begin{itemize}[leftmargin=*]
    \item \textbf{300-400B tokens.} 
    We want to guarantee our model will train on around 300-400B tokens of data. This is in the upper range for models in the size range we are pursuing, ensuring that low-resource languages will not be allocated too few tokens. Using the $C=6ND$ approximation \cite{kaplan2020scaling}, with $C=4,500$ PF-days and $D=300$-400B tokens, this constrains the model size to be around 160-200B parameters.
    \item \textbf{70-80 layers.} 
    From \citet{levine2020limits} and the size constraint above, we estimate that our model should have between 70 and 80 layers. 
    \item \textbf{Maximum throughput.}
    Finally, we want the final architecture to have as high of a throughput per GPU as possible, as more compute will translate directly into longer pretraining and thus a better model. Engineering constraints also come into light here: wide shallow models are typically easier to parallelize across nodes, up to a point where excessive tensor paralellism becomes necessary due to memory constraints.
\end{itemize}

We detail in Table \ref{tab:sota_models} the architectures of current state-of-the-art 100B+ models. From these guidelines, we benchmark 20 model configurations, detailed in Appendix~\ref{sec:arch_details}. Among these configurations, we select three of particular interest, outlined in Table~\ref{tab:final_configs}. They best fit our guidelines above, and offer high throughput, maximizing our training budget.

We discard configuration (1), as its attention heads are much larger than other models in the literature. Configuration (3) is shallower than recommended by \citet{levine2020limits}, but delivers 3\% higher throughput compared to (2). Thus, we choose configuration (3) and its better throughput, and because a shallower model is easier to deal with at inference time by introducing less latency.

\section{Limitations}

\paragraph{Optimal scaling.} Concurrent to this work,  \citet{hoffmann2022training} identified more optimal scaling laws. For our compute budget, they would suggest a 50B parameters model trained for a trillion tokens. Interestingly, even in hindsight, it would have been difficult to follow this recommendation as we would have been limited by the limited availability of high-quality multilingual data and by the size of the BigScience training dataset, ROOTS \cite{roots}. Note that our Figure \ref{fig:scaling} reproduces \citet{kaplan2020scaling} as we did not account for the learning rate schedule as suggested by \citet{hoffmann2022training}.

\paragraph{Other hyperparameters.} 
In this work we have focused on a subset of the available hyperparameter space of large language models. We have investigated architecture decisions around positional embeddings, activation functions and the embedding norm. Alternative attention mechanisms \cite{tay2020long} or optimizers are examples of other dimensions that could be investigated, potentially leading to improved models.

\paragraph{Efficient fine-tuning.} 
Our study is focused on zero-shot use and does not consider efficient fine-tuning \cite{lester2021power, zaken2021bitfit}, which is quite relevant for large language models, and which may lead to different conclusions. 
 
\section{Conclusion}

Seeking to establish the best possible model architecture that can be accommodated within a fixed 1,000,000 GPU-hours compute budget, we have presented an extensive study on principled modeling decisions for large language models.

First, we have found that complimenting Common Crawl data with high-quality cross-domain curated data can boost zero-shot generalization, validating previous suggestions \cite{rossettnlg, gao2020pile}. Through an ablation study, we have identified ALiBi as the position embedding of choice, confirmed the potential of SwiGLU, and highlighted that stabilizing techniques such as embedding normalization sometimes come at the expense of zero-shot generalization. Exploring multilinguality, we have found that multilingual models significantly underperform their monolingual counterparts on English zero-shot benchmarks, but that they can learn under-resourced languages along with larger ones if given enough scale. Finally, we identified a candidate architecture for BLOOM 176B, outlining the full reasoning behind every architectural parameter, including model shape. 

At variance with previous 100B+ models, such as GPT-3 \cite{brown2020gpt3} or Gopher \cite{rae2021scaling}, this project was conducted in the open, and resulted in a number of open-access artefacts. Notable similar projects conducted in parallel to this one include OPT \cite{zhang2022opt} and GLM \cite{zeng2022glm}, although they lacked the collaborative and massively multilingual components of this project.

We hope our work can help practitioners better understand modeling decisions, leading to better language models, and that this transparency will accelerate future similar work.

\subsection*{Acknowledgements}
This work was granted access to the HPC resources of Institut du d\'eveloppement et des ressources en informatique scientifique (IDRIS) du Centre national de la recherche scientifique (CNRS) under the allocation 2021-A0101012475 made by Grand \'equipement national de calcul intensif (GENCI). In particular, all the trainings ran on the Jean-Zay cluster of IDRIS, and we want to thank the IDRIS team for responsive support throughout the project, in particular R\'emi Lacroix.
Evaluations of GPT-3 models were provided in part by the Allen Institute for Artificial Intelligence. We thank Leo Gao for his expertise and advice on language model evaluation. 


\bibliography{custom}

\begin{thebibliography}{79}
\expandafter\ifx\csname natexlab\endcsname\relax\def\natexlab#1{#1}\fi

\bibitem[{Amini et~al.(2019)Amini, Gabriel, Lin, Koncel-Kedziorski, Choi, and
  Hajishirzi}]{amini2019mathqa}
Aida Amini, Saadia Gabriel, Shanchuan Lin, Rik Koncel-Kedziorski, Yejin Choi,
  and Hannaneh Hajishirzi. 2019.
\newblock \href {https://doi.org/10.18653/v1/N19-1245} {{M}ath{QA}: Towards
  interpretable math word problem solving with operation-based formalisms}.
\newblock In \emph{Proceedings of the 2019 Conference of the North {A}merican
  Chapter of the Association for Computational Linguistics: Human Language
  Technologies, Volume 1 (Long and Short Papers)}, pages 2357--2367,
  Minneapolis, Minnesota. Association for Computational Linguistics.

\bibitem[{Aroca-Ouellette et~al.(2021)Aroca-Ouellette, Paik, Roncone, and
  Kann}]{aroca-ouellette2021prost}
St{\'e}phane Aroca-Ouellette, Cory Paik, Alessandro Roncone, and Katharina
  Kann. 2021.
\newblock \href {https://aclanthology.org/2021.findings-acl.404} {{PROST}:
  {P}hysical reasoning about objects through space and time}.
\newblock In \emph{Findings of the Association for Computational Linguistics:
  ACL-IJCNLP 2021}, pages 4597--4608, Online. Association for Computational
  Linguistics.

\bibitem[{Ba et~al.(2016)Ba, Kiros, and Hinton}]{layernorm}
Jimmy~Lei Ba, Jamie~Ryan Kiros, and Geoffrey~E Hinton. 2016.
\newblock Layer normalization.
\newblock \emph{arXiv preprint arXiv:1607.06450}.

\bibitem[{Ben~Zhou and Roth(2019)}]{zhou2019mctaco}
Qiang~Ning Ben~Zhou, Daniel~Khashabi and Dan Roth. 2019.
\newblock “going on a vacation” takes longer than “going for a walk”: A
  study of temporal commonsense understanding.
\newblock In \emph{EMNLP}.

\bibitem[{Berant et~al.(2013)Berant, Chou, Frostig, and
  Liang}]{berant2013semantic}
Jonathan Berant, Andrew Chou, Roy Frostig, and Percy Liang. 2013.
\newblock Semantic parsing on freebase from question-answer pairs.
\newblock In \emph{Proceedings of the 2013 conference on empirical methods in
  natural language processing}, pages 1533--1544.

\bibitem[{Biderman(2021)}]{biderman2021nopos}
[@BlancheMinerva]~Stella Biderman. 2021.
\newblock \href
  {https://mobile.twitter.com/BlancheMinerva/status/1394089508723900422} {You:
  Gee stella, \#eleutherai sure hypes rotary embeddings a lot. are you sure
  that they're that good? me:}.
\newblock Twitter.

\bibitem[{{BigScience Workshop}(2022)}]{bigscience_workshop_2022}
{BigScience Workshop}. 2022.
\newblock \href {https://doi.org/10.57967/hf/0003} {Bloom (revision 4ab0472)}.

\bibitem[{Bisk et~al.(2020)Bisk, Zellers, Bras, Gao, and Choi}]{bisk2020piqa}
Yonatan Bisk, Rowan Zellers, Ronan~Le Bras, Jianfeng Gao, and Yejin Choi. 2020.
\newblock Piqa: Reasoning about physical commonsense in natural language.
\newblock In \emph{Thirty-Fourth AAAI Conference on Artificial Intelligence}.

\bibitem[{Black et~al.(2021)Black, Gao, Wang, Leahy, and Biderman}]{gpt-neo}
Sid Black, Leo Gao, Phil Wang, Connor Leahy, and Stella Biderman. 2021.
\newblock \href {https://doi.org/10.5281/zenodo.5297715} {{GPT-Neo: Large Scale
  Autoregressive Language Modeling with Mesh-Tensorflow}}.
\newblock {If you use this software, please cite it using these metadata.}

\bibitem[{Brown et~al.(2020)Brown, Mann, Ryder, Subbiah, Kaplan, Dhariwal,
  Neelakantan, Shyam, Sastry, Askell, Agarwal, Herbert-Voss, Krueger, Henighan,
  Child, Ramesh, Ziegler, Wu, Winter, Hesse, Chen, Sigler, Litwin, Gray, Chess,
  Clark, Berner, McCandlish, Radford, Sutskever, and Amodei}]{brown2020gpt3}
Tom Brown, Benjamin Mann, Nick Ryder, Melanie Subbiah, Jared~D Kaplan, Prafulla
  Dhariwal, Arvind Neelakantan, Pranav Shyam, Girish Sastry, Amanda Askell,
  Sandhini Agarwal, Ariel Herbert-Voss, Gretchen Krueger, Tom Henighan, Rewon
  Child, Aditya Ramesh, Daniel Ziegler, Jeffrey Wu, Clemens Winter, Chris
  Hesse, Mark Chen, Eric Sigler, Mateusz Litwin, Scott Gray, Benjamin Chess,
  Jack Clark, Christopher Berner, Sam McCandlish, Alec Radford, Ilya Sutskever,
  and Dario Amodei. 2020.
\newblock Language models are few-shot learners.
\newblock In \emph{Advances in Neural Information Processing Systems},
  volume~33, pages 1877--1901.

\bibitem[{Child et~al.(2019)Child, Gray, Radford, and Sutskever}]{sparse}
Rewon Child, Scott Gray, Alec Radford, and Ilya Sutskever. 2019.
\newblock Generating long sequences with sparse transformers.
\newblock \emph{URL https://openai.com/blog/sparse-transformers}.

\bibitem[{Clark et~al.(2019)Clark, Lee, Chang, Kwiatkowski, Collins, and
  Toutanova}]{clark2019boolq}
Christopher Clark, Kenton Lee, Ming-Wei Chang, Tom Kwiatkowski, Michael
  Collins, and Kristina Toutanova. 2019.
\newblock Boolq: Exploring the surprising difficulty of natural yes/no
  questions.
\newblock In \emph{NAACL}.

\bibitem[{Clark et~al.(2018)Clark, Cowhey, Etzioni, Khot, Sabharwal, Schoenick,
  and Tafjord}]{clark2018arc}
Peter Clark, Isaac Cowhey, Oren Etzioni, Tushar Khot, Ashish Sabharwal, Carissa
  Schoenick, and Oyvind Tafjord. 2018.
\newblock \href {http://arxiv.org/abs/1803.05457} {Think you have solved
  question answering? try arc, the {AI2} reasoning challenge}.
\newblock \emph{CoRR}, abs/1803.05457.

\bibitem[{Conneau et~al.(2018)Conneau, Rinott, Lample, Williams, Bowman,
  Schwenk, and Stoyanov}]{conneau2018xnli}
Alexis Conneau, Ruty Rinott, Guillaume Lample, Adina Williams, Samuel~R.
  Bowman, Holger Schwenk, and Veselin Stoyanov. 2018.
\newblock Xnli: Evaluating cross-lingual sentence representations.
\newblock In \emph{Proceedings of the 2018 Conference on Empirical Methods in
  Natural Language Processing}. Association for Computational Linguistics.

\bibitem[{Dagan et~al.(2005)Dagan, Glickman, and Magnini}]{dagan2005rte}
Ido Dagan, Oren Glickman, and Bernardo Magnini. 2005.
\newblock The pascal recognising textual entailment challenge.
\newblock In \emph{Machine Learning Challenges Workshop}, pages 177--190.
  Springer.

\bibitem[{Dauphin et~al.(2016)Dauphin, Fan, Auli, and
  Grangier}]{dauphin2016glu}
Yann~N. Dauphin, Angela Fan, Michael Auli, and David Grangier. 2016.
\newblock \href {http://arxiv.org/abs/1612.08083} {Language modeling with gated
  convolutional networks}.
\newblock \emph{CoRR}, abs/1612.08083.

\bibitem[{Dettmers et~al.(2021)Dettmers, Lewis, Shleifer, and
  Zettlemoyer}]{bitsandbytes}
Tim Dettmers, Mike Lewis, Sam Shleifer, and Luke Zettlemoyer. 2021.
\newblock \href {http://arxiv.org/abs/2110.02861} {8-bit optimizers via
  block-wise quantization}.

\bibitem[{Dolan and Brockett(2005)}]{dolan2016mrpc}
William~B Dolan and Chris Brockett. 2005.
\newblock Automatically constructing a corpus of sentential paraphrases.
\newblock In \emph{Proceedings of the Third International Workshop on
  Paraphrasing (IWP2005)}.

\bibitem[{Dong et~al.(2019)Dong, Yang, Wang, Wei, Liu, Wang, Gao, Zhou, and
  Hon}]{dong2019unified}
Li~Dong, Nan Yang, Wenhui Wang, Furu Wei, Xiaodong Liu, Yu~Wang, Jianfeng Gao,
  Ming Zhou, and Hsiao-Wuen Hon. 2019.
\newblock Unified language model pre-training for natural language
  understanding and generation.
\newblock \emph{Advances in Neural Information Processing Systems}, 32.

\bibitem[{Ganguli et~al.(2022)Ganguli, Hernandez, Lovitt, DasSarma, Henighan,
  Jones, Joseph, Kernion, Mann, Askell et~al.}]{ganguli2022predictability}
Deep Ganguli, Danny Hernandez, Liane Lovitt, Nova DasSarma, Tom Henighan, Andy
  Jones, Nicholas Joseph, Jackson Kernion, Ben Mann, Amanda Askell, et~al.
  2022.
\newblock Predictability and surprise in large generative models.
\newblock \emph{arXiv preprint arXiv:2202.07785}.

\bibitem[{Gao(2021)}]{gaosize}
Leo Gao. 2021.
\newblock \href {https://blog.eleuther.ai/gpt3-model-sizes/} {On the sizes of
  openai api models}.

\bibitem[{Gao et~al.(2020)Gao, Biderman, Black, Golding, Hoppe, Foster, Phang,
  He, Thite, Nabeshima, Presser, and Leahy}]{gao2020pile}
Leo Gao, Stella Biderman, Sid Black, Laurence Golding, Travis Hoppe, Charles
  Foster, Jason Phang, Horace He, Anish Thite, Noa Nabeshima, Shawn Presser,
  and Connor Leahy. 2020.
\newblock \href {https://arxiv.org/abs/2101.00027} {{The Pile:} an {800GB}
  dataset of diverse text for language modeling}.
\newblock \emph{arXiv preprint arXiv:2101.00027}.

\bibitem[{Gao et~al.(2021)Gao, Tow, Biderman, Black, DiPofi, Foster, Golding,
  Hsu, McDonell, Muennighoff, Phang, Reynolds, Tang, Thite, Wang, Wang, and
  Zou}]{eval-harness}
Leo Gao, Jonathan Tow, Stella Biderman, Sid Black, Anthony DiPofi, Charles
  Foster, Laurence Golding, Jeffrey Hsu, Kyle McDonell, Niklas Muennighoff,
  Jason Phang, Laria Reynolds, Eric Tang, Anish Thite, Ben Wang, Kevin Wang,
  and Andy Zou. 2021.
\newblock \href {https://doi.org/10.5281/zenodo.5371628} {A framework for
  few-shot language model evaluation}.

\bibitem[{Gordon et~al.(2012)Gordon, Kozareva, and Roemmele}]{gordon2012copa}
Andrew Gordon, Zornitsa Kozareva, and Melissa Roemmele. 2012.
\newblock \href {https://aclanthology.org/S12-1052} {{S}em{E}val-2012 task 7:
  Choice of plausible alternatives: An evaluation of commonsense causal
  reasoning}.
\newblock In \emph{*{SEM} 2012: The First Joint Conference on Lexical and
  Computational Semantics {--} Volume 1: Proceedings of the main conference and
  the shared task, and Volume 2: Proceedings of the Sixth International
  Workshop on Semantic Evaluation ({S}em{E}val 2012)}, pages 394--398,
  Montr{\'e}al, Canada. Association for Computational Linguistics.

\bibitem[{Hendrycks and Gimpel(2016)}]{hendrycks2016gaussian}
Dan Hendrycks and Kevin Gimpel. 2016.
\newblock Gaussian error linear units (gelus).
\newblock \emph{arXiv preprint arXiv:1606.08415}.

\bibitem[{Hoffmann et~al.(2022)Hoffmann, Borgeaud, Mensch, Buchatskaya, Cai,
  Rutherford, Casas, Hendricks, Welbl, Clark et~al.}]{hoffmann2022training}
Jordan Hoffmann, Sebastian Borgeaud, Arthur Mensch, Elena Buchatskaya, Trevor
  Cai, Eliza Rutherford, Diego de~Las Casas, Lisa~Anne Hendricks, Johannes
  Welbl, Aidan Clark, et~al. 2022.
\newblock Training compute-optimal large language models.
\newblock \emph{arXiv preprint arXiv:2203.15556}.

\bibitem[{Hu et~al.(2020)Hu, Ruder, Siddhant, Neubig, Firat, and
  Johnson}]{Hu2020XTREMEAM}
Junjie Hu, Sebastian Ruder, Aditya Siddhant, Graham Neubig, Orhan Firat, and
  Melvin Johnson. 2020.
\newblock Xtreme: A massively multilingual multi-task benchmark for evaluating
  cross-lingual generalization.
\newblock \emph{ArXiv}, abs/2003.11080.

\bibitem[{Iyer et~al.(2017)Iyer, Dandekar, and Csernai}]{iyer2019qqp}
Shankar Iyer, Nikhil Dandekar, and Kornel Csernai. 2017.
\newblock \href
  {https://data.quora.com/First-Quora-Dataset-Release-Question-Pairs} {First
  quora dataset release: Question pairs}.

\bibitem[{Jin et~al.(2019)Jin, Dhingra, Liu, Cohen, and Lu}]{jin2019pubmedqa}
Qiao Jin, Bhuwan Dhingra, Zhengping Liu, William Cohen, and Xinghua Lu. 2019.
\newblock Pubmedqa: A dataset for biomedical research question answering.
\newblock In \emph{Proceedings of the 2019 Conference on Empirical Methods in
  Natural Language Processing and the 9th International Joint Conference on
  Natural Language Processing (EMNLP-IJCNLP)}, pages 2567--2577.

\bibitem[{Johannes~Welbl(2017)}]{welbl2017sciq}
Matt~Gardner Johannes~Welbl, Nelson F.~Liu. 2017.
\newblock Crowdsourcing multiple choice science questions.

\bibitem[{Joshi et~al.(2017)Joshi, Choi, Weld, and
  Zettlemoyer}]{joshi2017triviaqa}
Mandar Joshi, Eunsol Choi, Daniel~S. Weld, and Luke Zettlemoyer. 2017.
\newblock Triviaqa: A large scale distantly supervised challenge dataset for
  reading comprehension.
\newblock In \emph{Proceedings of the 55th Annual Meeting of the Association
  for Computational Linguistics}, Vancouver, Canada. Association for
  Computational Linguistics.

\bibitem[{Kaplan et~al.(2020)Kaplan, McCandlish, Henighan, Brown, Chess, Child,
  Gray, Radford, Wu, and Amodei}]{kaplan2020scaling}
Jared Kaplan, Sam McCandlish, Tom Henighan, Tom~B Brown, Benjamin Chess, Rewon
  Child, Scott Gray, Alec Radford, Jeffrey Wu, and Dario Amodei. 2020.
\newblock Scaling laws for neural language models.
\newblock \emph{arXiv preprint arXiv:2001.08361}.

\bibitem[{Khashabi et~al.(2018)Khashabi, Chaturvedi, Roth, Upadhyay, and
  Roth}]{kashabi2018multirc}
Daniel Khashabi, Snigdha Chaturvedi, Michael Roth, Shyam Upadhyay, and Dan
  Roth. 2018.
\newblock Looking beyond the surface:a challenge set for reading comprehension
  over multiple sentences.
\newblock In \emph{Proceedings of North American Chapter of the Association for
  Computational Linguistics (NAACL)}.

\bibitem[{Kim et~al.(2021)Kim, Kim, Lee, Lee, Kwak, Jeon, Park, Kim, Kim, Seo,
  Lee, Jeong, Lee, Kim, Ko, Kim, Park, Kim, Kang, Ryu, Yoo, Chang, Suh, In,
  Park, Kim, Kim, Jeong, Yeo, hyun Ham, Park, Lee, Kang, Kang, Ha, Park, and
  Sung}]{Kim2021WhatCC}
Boseop Kim, Hyoungseok Kim, Sang-Woo Lee, Gichang Lee, Donghyun Kwak,
  Dong~Hyeon Jeon, Sunghyun Park, Sungju Kim, Seonhoon Kim, Dong~Hyung Seo,
  Heungsub Lee, Minyoung Jeong, Sungjae Lee, Minsub Kim, SukHyun Ko, Seokhun
  Kim, Taeyong Park, Jinuk Kim, Soyoung Kang, Na-Hyeon Ryu, Kang~Min Yoo,
  Minsuk Chang, Soobin Suh, Sookyo In, Jinseong Park, Kyungduk Kim, Hiun Kim,
  Jisu Jeong, Yong~Goo Yeo, Dong hyun Ham, Do-Hyoung Park, Min~Young Lee,
  Jaewoo Kang, Inho Kang, Jung-Woo Ha, Woo~Chul Park, and Nako Sung. 2021.
\newblock What changes can large-scale language models bring? intensive study
  on hyperclova: Billions-scale korean generative pretrained transformers.
\newblock \emph{ArXiv}, abs/2109.04650.

\bibitem[{Kingma and Ba(2014)}]{kingma2014adam}
Diederik~P Kingma and Jimmy Ba. 2014.
\newblock Adam: A method for stochastic optimization.
\newblock \emph{arXiv preprint arXiv:1412.6980}.

\bibitem[{Lai et~al.(2017)Lai, Xie, Liu, Yang, and Hovy}]{lai2017large}
Guokun Lai, Qizhe Xie, Hanxiao Liu, Yiming Yang, and Eduard Hovy. 2017.
\newblock Race: Large-scale reading comprehension dataset from examinations.
\newblock \emph{arXiv preprint arXiv:1704.04683}.

\bibitem[{Lashgar et~al.(2013)Lashgar, Baniasadi, and
  Khonsari}]{Lashgar2013WarpSI}
Ahmad Lashgar, Amirali Baniasadi, and Ahmad Khonsari. 2013.
\newblock Warp size impact in gpus: large or small?
\newblock In \emph{GPGPU@ASPLOS}.

\bibitem[{Lauren{\c{c}}on et~al.(2022)Lauren{\c{c}}on, Saulnier, Wang, Akiki,
  del Moral, Scao, Werra, Mou, Ponferrada, Nguyen, Frohberg, {\v{S}}a{\v{s}}ko,
  Lhoest, McMillan-Major, Dupont, Biderman, Rogers, allal, Toni, Pistilli,
  Nguyen, Nikpoor, Masoud, Colombo, de~la Rosa, Villegas, Thrush, Longpre,
  Nagel, Weber, Mu{\~n}oz, Zhu, Strien, Alyafeai, Almubarak, Chien,
  Gonzalez-Dios, Soroa, Lo, Dey, Suarez, Gokaslan, Bose, Adelani, Phan, Tran,
  Yu, Pai, Chim, Lepercq, Ilic, Mitchell, Luccioni, and Jernite}]{roots}
Hugo Lauren{\c{c}}on, Lucile Saulnier, Thomas Wang, Christopher Akiki,
  Albert~Villanova del Moral, Teven~Le Scao, Leandro~Von Werra, Chenghao Mou,
  Eduardo~Gonz{\'a}lez Ponferrada, Huu Nguyen, J{\"o}rg Frohberg, Mario
  {\v{S}}a{\v{s}}ko, Quentin Lhoest, Angelina McMillan-Major, G{\'e}rard
  Dupont, Stella Biderman, Anna Rogers, Loubna~Ben allal, Francesco~De Toni,
  Giada Pistilli, Olivier Nguyen, Somaieh Nikpoor, Maraim Masoud, Pierre
  Colombo, Javier de~la Rosa, Paulo Villegas, Tristan Thrush, Shayne Longpre,
  Sebastian Nagel, Leon Weber, Manuel~Romero Mu{\~n}oz, Jian Zhu, Daniel~Van
  Strien, Zaid Alyafeai, Khalid Almubarak, Vu~Minh Chien, Itziar Gonzalez-Dios,
  Aitor Soroa, Kyle Lo, Manan Dey, Pedro~Ortiz Suarez, Aaron Gokaslan, Shamik
  Bose, David~Ifeoluwa Adelani, Long Phan, Hieu Tran, Ian Yu, Suhas Pai, Jenny
  Chim, Violette Lepercq, Suzana Ilic, Margaret Mitchell, Sasha Luccioni, and
  Yacine Jernite. 2022.
\newblock \href {https://openreview.net/forum?id=UoEw6KigkUn} {The bigscience
  {ROOTS} corpus: A 1.6{TB} composite multilingual dataset}.
\newblock In \emph{Thirty-sixth Conference on Neural Information Processing
  Systems Datasets and Benchmarks Track}.

\bibitem[{Lester et~al.(2021)Lester, Al-Rfou, and Constant}]{lester2021power}
Brian Lester, Rami Al-Rfou, and Noah Constant. 2021.
\newblock The power of scale for parameter-efficient prompt tuning.
\newblock \emph{arXiv preprint arXiv:2104.08691}.

\bibitem[{Levesque et~al.(2012)Levesque, Davis, and
  Morgenstern}]{levesque2012winograd}
Hector Levesque, Ernest Davis, and Leora Morgenstern. 2012.
\newblock The winograd schema challenge.
\newblock In \emph{Thirteenth International Conference on the Principles of
  Knowledge Representation and Reasoning}.

\bibitem[{Levine et~al.(2020)Levine, Wies, Sharir, Bata, and
  Shashua}]{levine2020limits}
Yoav Levine, Noam Wies, Or~Sharir, Hofit Bata, and Amnon Shashua. 2020.
\newblock Limits to depth efficiencies of self-attention.
\newblock \emph{Advances in Neural Information Processing Systems},
  33:22640--22651.

\bibitem[{Lieber et~al.(2021)Lieber, Sharir, Lenz, and Shoham}]{J1WhitePaper}
Opher Lieber, Or~Sharir, Barak Lenz, and Yoav Shoham. 2021.
\newblock Jurassic-1: Technical details and evaluation.
\newblock Technical report, AI21 Labs.

\bibitem[{Lin et~al.(2021)Lin, Mihaylov, Artetxe, Wang, Chen, Simig, Ott,
  Goyal, Bhosale, Du, Pasunuru, Shleifer, Koura, Chaudhary, O'Horo, Wang,
  Zettlemoyer, Kozareva, Diab, Stoyanov, and Li}]{XGLM}
Xi~Victoria Lin, Todor Mihaylov, Mikel Artetxe, Tianlu Wang, Shuohui Chen,
  Daniel Simig, Myle Ott, Naman Goyal, Shruti Bhosale, Jingfei Du, Ramakanth
  Pasunuru, Sam Shleifer, Punit~Singh Koura, Vishrav Chaudhary, Brian O'Horo,
  Jeff Wang, Luke Zettlemoyer, Zornitsa Kozareva, Mona Diab, Ves Stoyanov, and
  Xian Li. 2021.
\newblock Few-shot learning with multilingual language models.
\newblock \emph{ArXiv}, abs/2112.10668.

\bibitem[{Liu et~al.(2020)Liu, Cui, Liu, Huang, Wang, and
  Zhang}]{liu2020logiqa}
Jian Liu, Leyang Cui, Hanmeng Liu, Dandan Huang, Yile Wang, and Yue Zhang.
  2020.
\newblock \href {http://arxiv.org/abs/2007.08124} {Logiqa: {A} challenge
  dataset for machine reading comprehension with logical reasoning}.
\newblock \emph{CoRR}, abs/2007.08124.

\bibitem[{Liu et~al.(2018)Liu, Saleh, Pot, Goodrich, Sepassi, Kaiser, and
  Shazeer}]{liu2018generating}
Peter~J Liu, Mohammad Saleh, Etienne Pot, Ben Goodrich, Ryan Sepassi, Lukasz
  Kaiser, and Noam Shazeer. 2018.
\newblock Generating wikipedia by summarizing long sequences.
\newblock In \emph{International Conference on Learning Representations}.

\bibitem[{Mihaylov et~al.(2018)Mihaylov, Clark, Khot, and
  Sabharwal}]{mihaylov2press2021train018openbookqa}
Todor Mihaylov, Peter Clark, Tushar Khot, and Ashish Sabharwal. 2018.
\newblock Can a suit of armor conduct electricity? a new dataset for open book
  question answering.
\newblock In \emph{EMNLP}.

\bibitem[{Narayanan et~al.(2021)Narayanan, Shoeybi, Casper, LeGresley, Patwary,
  Korthikanti, Vainbrand, Kashinkunti, Bernauer, Catanzaro
  et~al.}]{narayanan2021efficient}
Deepak Narayanan, Mohammad Shoeybi, Jared Casper, Patrick LeGresley, Mostofa
  Patwary, Vijay Korthikanti, Dmitri Vainbrand, Prethvi Kashinkunti, Julie
  Bernauer, Bryan Catanzaro, et~al. 2021.
\newblock Efficient large-scale language model training on gpu clusters using
  megatron-lm.
\newblock In \emph{Proceedings of the International Conference for High
  Performance Computing, Networking, Storage and Analysis}, pages 1--15.

\bibitem[{{Ortiz Su{\'a}rez} et~al.(2019){Ortiz Su{\'a}rez}, Sagot, and
  Romary}]{ortiz2019oscar}
Pedro~Javier {Ortiz Su{\'a}rez}, Beno{\^i}t Sagot, and Laurent Romary. 2019.
\newblock \href {https://doi.org/10.14618/ids-pub-9021} {Asynchronous pipelines
  for processing huge corpora on medium to low resource infrastructures}.
\newblock Proceedings of the Workshop on Challenges in the Management of Large
  Corpora (CMLC-7) 2019. Cardiff, 22nd July 2019, pages 9 -- 16, Mannheim.
  Leibniz-Institut f{\"u}r Deutsche Sprache.

\bibitem[{Paperno et~al.(2016)Paperno, Kruszewski, Lazaridou, Pham, Bernardi,
  Pezzelle, Baroni, Boleda, and Fern{\'a}ndez}]{paperno2016lambada}
Denis Paperno, Germ{\'a}n Kruszewski, Angeliki Lazaridou, Ngoc~Quan Pham,
  Raffaella Bernardi, Sandro Pezzelle, Marco Baroni, Gemma Boleda, and Raquel
  Fern{\'a}ndez. 2016.
\newblock \href {https://doi.org/10.18653/v1/P16-1144} {The {LAMBADA} dataset:
  Word prediction requiring a broad discourse context}.
\newblock In \emph{Proceedings of the 54th Annual Meeting of the Association
  for Computational Linguistics (Volume 1: Long Papers)}, pages 1525--1534,
  Berlin, Germany. Association for Computational Linguistics.

\bibitem[{Pilehvar and os{'{e} } Camacho{-}Collados(2018)}]{pilehavar2018wic}
Mohammad~Taher Pilehvar and os{'{e} } Camacho{-}Collados. 2018.
\newblock \href {http://arxiv.org/abs/1808.09121} {Wic: 10, 000 example pairs
  for evaluating context-sensitive representations}.
\newblock \emph{CoRR}, abs/1808.09121.

\bibitem[{Press et~al.(2022)Press, Smith, and Lewis}]{press2021alibi}
Ofir Press, Noah Smith, and Mike Lewis. 2022.
\newblock \href {https://openreview.net/forum?id=R8sQPpGCv0} {Train short, test
  long: Attention with linear biases enables input length extrapolation}.
\newblock In \emph{International Conference on Learning Representations}.

\bibitem[{Press and Wolf(2017)}]{tying}
Ofir Press and Lior Wolf. 2017.
\newblock \href {https://aclanthology.org/E17-2025} {Using the output embedding
  to improve language models}.
\newblock In \emph{Proceedings of the 15th Conference of the {E}uropean Chapter
  of the Association for Computational Linguistics: Volume 2, Short Papers},
  pages 157--163, Valencia, Spain. Association for Computational Linguistics.

\bibitem[{Radford et~al.(2019)Radford, Wu, Child, Luan, Amodei, and
  Sutskever}]{radford2019language}
Alec Radford, Jeff Wu, Rewon Child, David Luan, Dario Amodei, and Ilya
  Sutskever. 2019.
\newblock Language models are unsupervised multitask learners.

\bibitem[{Rae et~al.(2021)Rae, Borgeaud, Cai, Millican, Hoffmann, Song,
  Aslanides, Henderson, Ring, Young et~al.}]{rae2021scaling}
Jack~W Rae, Sebastian Borgeaud, Trevor Cai, Katie Millican, Jordan Hoffmann,
  Francis Song, John Aslanides, Sarah Henderson, Roman Ring, Susannah Young,
  et~al. 2021.
\newblock Scaling language models: Methods, analysis \& insights from training
  gopher.
\newblock \emph{arXiv preprint arXiv:2112.11446}.

\bibitem[{Raffel et~al.(2019)Raffel, Shazeer, Roberts, Lee, Narang, Matena,
  Zhou, Li, and Liu}]{raffel2019t5}
Colin Raffel, Noam Shazeer, Adam Roberts, Katherine Lee, Sharan Narang, Michael
  Matena, Yanqi Zhou, Wei Li, and Peter~J. Liu. 2019.
\newblock \href {http://arxiv.org/abs/1910.10683} {Exploring the limits of
  transfer learning with a unified text-to-text transformer}.
\newblock \emph{CoRR}, abs/1910.10683.

\bibitem[{Rajpurkar et~al.(2016)Rajpurkar, Zhang, Lopyrev, and
  Liang}]{rajpurkar2016squad}
Pranav Rajpurkar, Jian Zhang, Konstantin Lopyrev, and Percy Liang. 2016.
\newblock Squad: 100,000+ questions for machine comprehension of text.
\newblock \emph{arXiv preprint arXiv:1606.05250}.

\bibitem[{Ramachandran et~al.(2017)Ramachandran, Zoph, and
  Le}]{ramachandran2017searching}
Prajit Ramachandran, Barret Zoph, and Quoc~V Le. 2017.
\newblock Searching for activation functions.
\newblock \emph{arXiv preprint arXiv:1710.05941}.

\bibitem[{Rosset(2020)}]{rossettnlg}
Corby Rosset. 2020.
\newblock \href
  {https://www.microsoft.com/en-us/research/blog/turing-nlg-a-17-billion-parameter-language-model-by-microsoft/}
  {Turing-nlg: A 17-billion-parameter language model by microsoft}.

\bibitem[{Sakaguchi et~al.(2019)Sakaguchi, Bras, Bhagavatula, and
  Choi}]{sakaguchi2019winogrande}
Keisuke Sakaguchi, Ronan~Le Bras, Chandra Bhagavatula, and Yejin Choi. 2019.
\newblock Winogrande: An adversarial winograd schema challenge at scale.
\newblock \emph{arXiv preprint arXiv:1907.10641}.

\bibitem[{Sanh et~al.(2021)Sanh, Webson, Raffel, Bach, Sutawika, Alyafeai,
  Chaffin, Stiegler, Scao, Raja, Dey, BARI, Xu, Thakker, Sharma, Szczechla,
  Kim, Chhablani, Nayak, Datta, Chang, Jiang, Wang, Manica, Shen, Yong, Pandey,
  Bawden, Wang, Neeraj, Rozen, Sharma, Santilli, F{\'e}vry, Fries, Teehan,
  Biderman, Gao, Bers, Wolf, and Rush}]{Sanh2021MultitaskPT}
Victor Sanh, Albert Webson, Colin Raffel, Stephen~H. Bach, Lintang~A. Sutawika,
  Zaid Alyafeai, Antoine Chaffin, Arnaud Stiegler, Teven~Le Scao, Arun Raja,
  Manan Dey, M~SAIFUL BARI, Canwen Xu, Urmish Thakker, Shanya~Sharma Sharma,
  Eliza Szczechla, Taewoon Kim, Gunjan Chhablani, Nihal~V. Nayak, Debajyoti
  Datta, Jonathan Chang, Mike Tian-Jian Jiang, Han Wang, Matteo Manica, Sheng
  Shen, Zheng~Xin Yong, Harshit Pandey, Rachel Bawden, Thomas Wang, Trishala
  Neeraj, Jos Rozen, Abheesht Sharma, Andrea Santilli, Thibault F{\'e}vry,
  Jason~Alan Fries, Ryan Teehan, Stella~Rose Biderman, Leo Gao, T.~G.~Owe Bers,
  Thomas Wolf, and Alexander~M. Rush. 2021.
\newblock Multitask prompted training enables zero-shot task generalization.
\newblock \emph{ArXiv}, abs/2110.08207.

\bibitem[{Shazeer(2020)}]{shazeer2020swiglu}
Noam Shazeer. 2020.
\newblock \href {http://arxiv.org/abs/2002.05202} {Glu variants improve
  transformer}.

\bibitem[{Smith et~al.(2022)Smith, Patwary, Norick, LeGresley, Rajbhandari,
  Casper, Liu, Prabhumoye, Zerveas, Korthikanti et~al.}]{smith2022using}
Shaden Smith, Mostofa Patwary, Brandon Norick, Patrick LeGresley, Samyam
  Rajbhandari, Jared Casper, Zhun Liu, Shrimai Prabhumoye, George Zerveas,
  Vijay Korthikanti, et~al. 2022.
\newblock Using deepspeed and megatron to train megatron-turing nlg 530b, a
  large-scale generative language model.
\newblock \emph{arXiv preprint arXiv:2201.11990}.

\bibitem[{Socher et~al.(2013)Socher, Perelygin, Wu, Chuang, Manning, Ng, and
  Potts}]{socher2013sst}
Richard Socher, Alex Perelygin, Jean Wu, Jason Chuang, Christopher~D Manning,
  Andrew Ng, and Christopher Potts. 2013.
\newblock Recursive deep models for semantic compositionality over a sentiment
  treebank.
\newblock In \emph{Proceedings of the 2013 conference on empirical methods in
  natural language processing}, pages 1631--1642.

\bibitem[{Su et~al.(2021)Su, Lu, Pan, Wen, and Liu}]{su2021roformer}
Jianlin Su, Yu~Lu, Shengfeng Pan, Bo~Wen, and Yunfeng Liu. 2021.
\newblock \href {https://arxiv.org/abs/2104.09864} {Roformer: Enhanced
  transformer with rotary position embedding}.
\newblock \emph{arXiv preprint arXiv:2104.09864}.

\bibitem[{Tay et~al.(2020)Tay, Dehghani, Abnar, Shen, Bahri, Pham, Rao, Yang,
  Ruder, and Metzler}]{tay2020long}
Yi~Tay, Mostafa Dehghani, Samira Abnar, Yikang Shen, Dara Bahri, Philip Pham,
  Jinfeng Rao, Liu Yang, Sebastian Ruder, and Donald Metzler. 2020.
\newblock Long range arena: A benchmark for efficient transformers.
\newblock \emph{arXiv preprint arXiv:2011.04006}.

\bibitem[{Tay et~al.(2021)Tay, Dehghani, Rao, Fedus, Abnar, Chung, Narang,
  Yogatama, Vaswani, and Metzler}]{Tay2021ScaleEI}
Yi~Tay, Mostafa Dehghani, Jinfeng Rao, William Fedus, Samira Abnar, Hyung~Won
  Chung, Sharan Narang, Dani Yogatama, Ashish Vaswani, and Donald Metzler.
  2021.
\newblock Scale efficiently: Insights from pre-training and fine-tuning
  transformers.
\newblock \emph{ArXiv}, abs/2109.10686.

\bibitem[{Thoppilan et~al.(2022)Thoppilan, De~Freitas, Hall, Shazeer,
  Kulshreshtha, Cheng, Jin, Bos, Baker, Du et~al.}]{thoppilan2022lamda}
Romal Thoppilan, Daniel De~Freitas, Jamie Hall, Noam Shazeer, Apoorv
  Kulshreshtha, Heng-Tze Cheng, Alicia Jin, Taylor Bos, Leslie Baker, Yu~Du,
  et~al. 2022.
\newblock Lamda: Language models for dialog applications.
\newblock \emph{arXiv preprint arXiv:2201.08239}.

\bibitem[{Vaswani et~al.(2017)Vaswani, Shazeer, Parmar, Uszkoreit, Jones,
  Gomez, Kaiser, and Polosukhin}]{vaswani2017attention}
Ashish Vaswani, Noam Shazeer, Niki Parmar, Jakob Uszkoreit, Llion Jones,
  Aidan~N Gomez, {\L}ukasz Kaiser, and Illia Polosukhin. 2017.
\newblock Attention is all you need.
\newblock In \emph{Advances in neural information processing systems}, pages
  5998--6008.

\bibitem[{Wang et~al.(2019)Wang, Singh, Michael, Hill, Levy, and
  Bowman}]{wang2019glue}
Alex Wang, Amanpreet Singh, Julian Michael, Felix Hill, Omer Levy, and
  Samuel~R. Bowman. 2019.
\newblock {GLUE}: A multi-task benchmark and analysis platform for natural
  language understanding.
\newblock In the Proceedings of ICLR.

\bibitem[{Wang et~al.(2022)Wang, Roberts, Hesslow, Scao, Chung, Beltagy,
  Launay, and Raffel}]{wang2022language}
Thomas Wang, Adam Roberts, Daniel Hesslow, Teven~Le Scao, Hyung~Won Chung,
  Iz~Beltagy, Julien Launay, and Colin Raffel. 2022.
\newblock What language model architecture and pretraining objective work best
  for zero-shot generalization?
\newblock \emph{arXiv preprint arXiv:2204.05832}.

\bibitem[{Wei et~al.(2021)Wei, Bosma, Zhao, Guu, Yu, Lester, Du, Dai, and
  Le}]{wei2021finetuned}
Jason Wei, Maarten Bosma, Vincent~Y Zhao, Kelvin Guu, Adams~Wei Yu, Brian
  Lester, Nan Du, Andrew~M Dai, and Quoc~V Le. 2021.
\newblock Finetuned language models are zero-shot learners.
\newblock \emph{arXiv preprint arXiv:2109.01652}.

\bibitem[{Wu et~al.(2021)Wu, Zhao, Yu, Zhang, Shen, Liu, Li, Zhu, Luo, Xu
  et~al.}]{wu2021yuan}
Shaohua Wu, Xudong Zhao, Tong Yu, Rongguo Zhang, Chong Shen, Hongli Liu, Feng
  Li, Hong Zhu, Jiangang Luo, Liang Xu, et~al. 2021.
\newblock Yuan 1.0: Large-scale pre-trained language model in zero-shot and
  few-shot learning.
\newblock \emph{arXiv preprint arXiv:2110.04725}.

\bibitem[{Xue et~al.(2020)Xue, Constant, Roberts, Kale, Al-Rfou, Siddhant,
  Barua, and Raffel}]{mT5}
Linting Xue, Noah Constant, Adam Roberts, Mihir Kale, Rami Al-Rfou, Aditya
  Siddhant, Aditya Barua, and Colin Raffel. 2020.
\newblock mt5: A massively multilingual pre-trained text-to-text transformer.
\newblock \emph{arXiv preprint arXiv:2010.11934}.

\bibitem[{Xue et~al.(2021)Xue, Constant, Roberts, Kale, Al-Rfou, Siddhant,
  Barua, and Raffel}]{Xue2021mT5AM}
Linting Xue, Noah Constant, Adam Roberts, Mihir Kale, Rami Al-Rfou, Aditya
  Siddhant, Aditya Barua, and Colin Raffel. 2021.
\newblock mt5: A massively multilingual pre-trained text-to-text transformer.
\newblock In \emph{NAACL}.

\bibitem[{Zaken et~al.(2021)Zaken, Ravfogel, and Goldberg}]{zaken2021bitfit}
Elad~Ben Zaken, Shauli Ravfogel, and Yoav Goldberg. 2021.
\newblock Bitfit: Simple parameter-efficient fine-tuning for transformer-based
  masked language-models.
\newblock \emph{arXiv preprint arXiv:2106.10199}.

\bibitem[{Zellers et~al.(2019)Zellers, Holtzman, Bisk, Farhadi, and
  Choi}]{zellers2019hellaswag}
Rowan Zellers, Ari Holtzman, Yonatan Bisk, Ali Farhadi, and Yejin Choi. 2019.
\newblock \href {https://doi.org/10.18653/v1/P19-1472} {{H}ella{S}wag: Can a
  machine really finish your sentence?}
\newblock In \emph{Proceedings of the 57th Annual Meeting of the Association
  for Computational Linguistics}, pages 4791--4800, Florence, Italy.
  Association for Computational Linguistics.

\bibitem[{Zeng et~al.(2022)Zeng, Liu, Du, Wang, Lai, Ding, Yang, Xu, Zheng, Xia
  et~al.}]{zeng2022glm}
Aohan Zeng, Xiao Liu, Zhengxiao Du, Zihan Wang, Hanyu Lai, Ming Ding, Zhuoyi
  Yang, Yifan Xu, Wendi Zheng, Xiao Xia, et~al. 2022.
\newblock Glm-130b: An open bilingual pre-trained model.
\newblock \emph{arXiv preprint arXiv:2210.02414}.

\bibitem[{Zeng et~al.(2021)Zeng, Ren, Su, Wang, Liao, Wang, Jiang, Yang, Wang,
  Zhang et~al.}]{zeng2021pangu}
Wei Zeng, Xiaozhe Ren, Teng Su, Hui Wang, Yi~Liao, Zhiwei Wang, Xin Jiang,
  ZhenZhang Yang, Kaisheng Wang, Xiaoda Zhang, et~al. 2021.
\newblock Pangu-$\alpha$: Large-scale autoregressive pretrained chinese
  language models with auto-parallel computation.
\newblock \emph{arXiv preprint arXiv:2104.12369}.

\bibitem[{Zhang et~al.(2022)Zhang, Roller, Goyal, Artetxe, Chen, Chen, Dewan,
  Diab, Li, Lin et~al.}]{zhang2022opt}
Susan Zhang, Stephen Roller, Naman Goyal, Mikel Artetxe, Moya Chen, Shuohui
  Chen, Christopher Dewan, Mona Diab, Xian Li, Xi~Victoria Lin, et~al. 2022.
\newblock Opt: Open pre-trained transformer language models.
\newblock \emph{arXiv preprint arXiv:2205.01068}.

\end{thebibliography}
\bibliographystyle{acl_natbib}

\newpage

\appendix
\onecolumn

\section{Open artefacts: models, code, and logs}
\label{sec:artefacts}
We make public all artefacts produced as part of this work:
\begin{itemize}
    \item \textbf{Models.} All trained models are centralized at \url{https://huggingface.co/bigscience};
    \item \textbf{Code.} All code is available at \url{https://github.com/bigscience-workshop/Megatron-DeepSpeed/tree/main/megatron};
    \item \textbf{Discussions and logbook.} The notes from the weekly meetings of our working group are made available at \url{https://docs.google.com/document/d/1qbIkhd6bvbOsJOWXL7SfKQ0jey3MWQYQb_SshqH1LII/}.
\end{itemize}

\section{Multilingual scaling laws}
\label{sec:multilingualscalinglaws}

\begin{table*}[h]
\label{tab:multilingualscalinglaws}
\begin{center}
\begin{tabular}{cccc}
\toprule
Language & Proportion [\%] & $\alpha_c$ & $C_m$ \\
\midrule

Arabic & 4.6 & 0.057 & 1.16 \\
Catalan & 1.1 & 0.057 & 1.11\\
Code & 10.8 & 0.054 & 0.94\\
English & 30.0 & 0.051 & 1.08 \\
Spanish & 10.8 & 0.050 & 1.01 \\
Basque & 0.15 & 0.069 & 1.28 \\
French & 12.9 & 0.047 & 1.06\\
Indonesian & 1.2 & 0.051 & 1.14\\
Assamese & 0.01 & 0.051 & 1.31\\
Bengali & 0.5 & 0.037 & 1.15\\
Gujarati & 0.04 & 0.051 & 1.30\\
Hindi & 0.7 & 0.045 & 1.14\\
Kannada & 0.06 & 0.046 & 1.26\\
Malayalam & 0.1 & 0.044 & 1.17\\
Marathi & 0.05 & 0.046 & 1.23\\
Nepali & 0.07 & 0.055 & 1.25 \\
Odia & 0.04 & 0.044 & 1.25\\
Punjabi & 0.05 & 0.043 & 1.20\\
Tamil & 0.2 & 0.030 & 1.14\\
Telugu & 0.09 & 0.056 & 1.31\\
Urdu & 0.1 & 0.068 & 1.31\\
Niger-Congo (family) & 0.03 & 0.039 & 1.22\\
Portuguese & 4.9 & 0.049 & 1.05\\
Vietnamese & 2.7 & 0.053 & 1.08\\
Chinese (simplified) & 16.2 & 0.052 & 1.09\\
Chinese (traditionnal) & 0.05 & 0.050 & 1.15\\
\bottomrule

\end{tabular}
\caption{\textbf{Best scaling law fit per language.} We fit $\mathcal{L}(C) = C_m C^{-\alpha_c}$ to the runs reported in Figure \ref{fig:multilingualscaling}. But for a handful of languages which are poorly represented in the overall mixture (Basque, most of the Indic family, and Niger-Congo languages), scaling mostly different in offset $C_m$, not in exponent $\alpha_c$.}
\end{center}

\end{table*}

\newpage

\section{Evaluation details}
\label{sec:sup_eval}
\vfill
\begin{table*}[h]
\label{tab:sup_random-baselines}
\begin{center}
\begin{tiny}
\begin{tabular}{lllc}
\toprule
\multicolumn{2}{c}{\textbf{Task}} & \textbf{Type}           &  \textbf{Random baseline}                                \\ 
\midrule
ARC \citep{clark2018arc}      & Challenge & Natural Language Inference & 25.0       \\
             & Easy &     & 25.0                         \\
GLUE         & MRPC  \citep{dolan2016mrpc} & Paraphrase Identification      & 50.0                                                  \\
             & QQP \citep{iyer2019qqp} & Paraphrase Identification      & 50.0                                                             \\  
HellaSwag \citep{zellers2019hellaswag}    & & Sentence Completion           & 25.0                 \\
LAMBADA \citep{paperno2016lambada}      & & Sentence Completion       & 0.0                                        \\
LogiQA \citep{liu2020logiqa}      & & Multiple-Choice Question Answering           & 25.0                                               \\
MathQA \citep{amini2019mathqa}       & & Multiple-Choice Question Answering           & 20.1                                          \\
MC-TACO \citep{zhou2019mctaco} & & Multiple-Choice Question Answering & 36.2 \\
OpenBookQA \citep{mihaylov2press2021train018openbookqa}  & & Multiple-Choice Question Answering          & 25.0       \\
PIQA \citep{bisk2020piqa}         &  & Multiple-Choice Question Answering          & 50.0        \\
PROST  \citep{aroca-ouellette2021prost}        & & Multiple-Choice Question Answering          & 25.0                          \\
PudMedQA \citep{jin2019pubmedqa}     & & Multiple-Choice Question Answering          & 33.3                                         \\
QNLI \citep{rajpurkar2016squad,wang2019glue}         & & Sentence Completion           & 50.0                                         \\
Race \cite{lai2017large}        & & Closed-Book Question Answering         & 25.0                       \\
SciQ \citep{welbl2017sciq}         & & Multiple-Choice Question Answering          & 25.0                                          \\
SST \citep{socher2013sst}         & & Sentiment          & 50.0                                                \\
SuperGLUE    & Boolq \citep{clark2019boolq} & Multiple-Choice Question Answering     & 50.0                     \\
             & COPA \citep{gordon2012copa} & Sentence Completion      & 50.0 \\
             & MultiRC \citep{kashabi2018multirc} & Multiple-Choice Question Answering   & 5.8                                          \\
             & RTE \citep{dagan2005rte} & Natural Language Inference       & 50.0                     \\
             & WIC  \citep{pilehavar2018wic} & Word Sense Disambiguation       & 50.0                     \\
             & WSC \citep{levesque2012winograd} & Word Sense Disambiguation      & 50.0                     \\
TriviaQA \citep{joshi2017triviaqa}     & & Closed-Book Question Answering          & 0.0                      \\
WebQuestions \citep{berant2013semantic} & & Closed-Book Question Answering         & 0.0                        \\
Winogrande \citep{sakaguchi2019winogrande}   & & Coreference resolution           & 50.0             \\
WNLI \citep{sakaguchi2019winogrande}        & & Natural Language Inference         & 50.0      \\ \midrule
\textbf{EAI harness} & & & \textbf{33.3} \\
\bottomrule
\end{tabular}
\end{tiny}
\end{center}
\caption{\textbf{Evaluation tasks considered in the EAI harness and random baselines.}}
\vskip -0.1in
\end{table*}
\vfill
\newpage

\section{Architecture details}
\label{sec:arch_details}

\vfill

\begin{table}[h]
\vskip 0.15in
\begin{center}
\begin{small}
\centerline{\begin{tabular}{@{}cccccccccccc@{}}
\toprule
\multicolumn{5}{c}{\textbf{\sc{Architecture}}}                                                                                     & \multicolumn{4}{c}{\textbf{\sc{Parallelism}}}                                       & \multicolumn{3}{c}{\textbf{\sc{Performance}}}                \\ \midrule
\multicolumn{1}{l}{\textbf{Size}} & \textbf{Hidden dim.}    & \textbf{Layers}      & \multicolumn{2}{c}{\textbf{Attention heads}} & \textbf{Data}       & \textbf{Tensor}    & \textbf{Pipeline}   & \textbf{MBS}       & \textbf{Memory}    & \multicolumn{2}{c}{\textbf{Throughput}} \\
 {[Bparams.]}                      &                         &                      & num.                 & dim.                  &                     &                    &                     &                    & [GB]               & [s/iter.]            & [TFLOPs]           \\ \midrule
206                               & 14,336                  & 82                   & 128                  & 112                   & 8                   & 4                  & 12                  & 2                  & \ul{OOM}           &                      &                  \\ \midrule
203                               & 13,312                  & 94                   & 128                  & 104                   & 8                   & 4                  & 12                  & 2                  & 67                 & 124,1                & 146,1            \\ \midrule 
\multirow{4}{*}{195}              & \multirow{4}{*}{12,288} & \multirow{4}{*}{106} & 128                  & 96                    & \multirow{4}{*}{8}  & \multirow{4}{*}{4} & \multirow{4}{*}{12} & 2                  & 67                 & 121,4                & 143,7            \\
                                  &                         &                      & \multirow{2}{*}{96}  & 128                   &                     &                    &                     & 4                  & 79                 & 120,3                & 145,0            \\
                                  &                         &                      &                      & 128                   &                     &                    &                     & \multirow{2}{*}{2} & 65                 & 118,8                & 146,9            \\
                                  &                         &                      & 64                   & 192                   &                     &                    &                     &                    & 67                 & 116,5                & 149,8            \\\midrule
\multirow{4}{*}{184}              & \multirow{4}{*}{12,288} & \multirow{4}{*}{100} & \multirow{4}{*}{64}  & \multirow{4}{*}{192}  & \multirow{2}{*}{16} & \multirow{2}{*}{4} & \multirow{4}{*}{6}  & 2                  & \ul{OOM}           &                      &                  \\
                                  &                         &                      &                      &                       &                     &                    &                     & 1                  & \ul{OOM}           &                      &                  \\
                                  &                         &                      &                      &                       & \multirow{2}{*}{8}  & \multirow{2}{*}{8} &                     & 4                  & 72                 & 121,0                & 136,2            \\
                                  &                         &                      &                      &                       &                     &                    &                     & 2                  & 61                 & 140,0                & 117,9            \\\midrule
\multirow{5}{*}{178}              & \multirow{5}{*}{13,312} & \multirow{5}{*}{82}  & 128                  & 104                   & \multirow{3}{*}{8}  & \multirow{3}{*}{4} & \multirow{5}{*}{12} & \multirow{2}{*}{2} & 60                 & 108,8                & 145,7            \\
                                  &                         &                      & 104                  & 128                   &                     &                    &                     &                    & 62                 & 123,7                & 128,1            \\
                                  &                         &                      & \multirow{3}{*}{64}  & \multirow{3}{*}{208}  &                     &                    &                     & \multirow{2}{*}{4} & 74                 & 104,8                & 151,2            \\
                                  &                         &                      &                      &                       & 4                   & 8                  &                     &                    & 52                 & 111,8                & 141,8            \\
                                  &                         &                      &                      &                       & 8                   & 4                  &                     & 2                  & 63                 & 104,5                & 151,7            \\\midrule
\multirow{5}{*}{176}              & \multirow{5}{*}{14,336} & \multirow{5}{*}{70}  & 128                  & 112                   & \multirow{4}{*}{8}  & \multirow{4}{*}{4} & \multirow{4}{*}{12} & \multirow{2}{*}{2} & 60                 & 105,9                & 148,1            \\
                                  &                         &                      & 112                  & 128                   &                     &                    &                     &                    & 59                 & 104,5                & 150,1            \\
                                  &                         &                      & \multirow{3}{*}{64}  & \multirow{3}{*}{224}  &                     &                    &                     & 4                  & 73                 & 102,3                & 153,3            \\
                                  &                         &                      &                      &                       &                     &                    &                     & \multirow{2}{*}{2} & 59                 & 102,0                & 153,7            \\
                                  &                         &                      &                      &                       & 4                   & 8                  & 12                  &                    & 40                 & 121,6                & 128,9            \\ \bottomrule

\end{tabular}}
\end{small}
\end{center}
\caption{\textbf{Throughput and memory usage of considered models sizes.} Note that pipeline parallelism here considers equal "slots" for embeddings and Transformer layers. This is important to optimize pipeline use, as our multilingual embeddings are quite large (250k).}
\label{tab:sup_throughput}
\end{table}

\vfill

\newpage
\section{All Results}
\label{sec:all_results}

\begin{table}[h]
\vskip 0.15in
\begin{center}
\begin{small}
\centerline{\begin{tabular}{lllllllll}
\toprule
Ablation       & Dataset  & Embedding & Activation & Embedding Norm & Parameters & 112GT   & 250GT   & 300GT   \\ \midrule
Embeddings     & OSCAR    & Learned   & GELU       & No             & 1.3B       & 41.71 &       &       \\
Embeddings     & OSCAR    & None      & GELU       & No             & 1.3B       & 41.23 &       &       \\
Embeddings     & OSCAR    & Rotary    & GELU       & No             & 1.3B       & 41.46 &       &       \\
Embeddings     & OSCAR    & ALiBi     & GELU       & No             & 1.3B       & 43.70 &       &       \\\midrule
Dataset        & The Pile & Learned   & GELU       & No             & 1.3B       & 42.79 & 43.12 & 43.46 \\
Dataset        & C4       & Learned   & GELU       & No             & 1.3B       & 42.77 &       &       \\
Dataset        & OSCAR    & Learned   & GELU       & No             & 1.3B       & 42.79 &       &       \\\midrule
Activation     & The Pile & Learned   & GELU       & No             & 1.3B       & 42.79 &       &       \\
Activation     & The Pile & Learned   & SwiGLU     & No             & 1.3B       & 42.95 &       &       \\\midrule
Embedding Norm & The Pile & Learned   & GELU       & No             & 1.3B       & 42.79 & 43.12 & 43.46 \\
Embedding Norm & The Pile & Learned   & GELU       & Yes            & 1.3B       &       &       & 42.24 \\\midrule
Multilinguality             & OSCAR-ML & Learned   & GELU       & No             & 1.3B       & 38.55 &       &       \\
Multilinguality             & OSCAR    & Learned   & GELU       & No             & 1.3B       & 41.72 &       &       \\\midrule
Scale           & OSCAR    & Learned   & GELU       & No             & 1.3B       & 41.72 &       &       \\
Scale           & OSCAR    & Learned   & GELU       & No             & 13B        &       &       & 47.09\\
\bottomrule
\end{tabular}}
\end{small}
\end{center}
\caption{\textbf{Summary of all results obtained in this study}. The final three columns indicate the average EAI Harness results at across different billion tokens trained. Some rows are duplicated for ease of reading.}

\label{tab:all_results}
\end{table}

\begin{sidewaystable}
\centering
\begin{tiny}
\begin{tabular}{lllllllllllllllllll}
Public Name                 & ~         & ~                       & OpenAI:
  babbage & Openai:
  curie & gpt-neo
  1.3B & ~       & ~       & ~        & ~        & ~        & ~        & ~        & ~       & ~        & ~       & ~       & ~       & ~         \\
Dataset                     & ~         & ~                       & ~                 & ~               & ~              & C4      & OSCAR   & The Pile & The Pile & The Pile & The Pile & The Pile & OSCAR   & The Pile & OSCAR   & OSCAR   & OSCAR   & OSCAR-ML  \\
Embeddings                  & ~         & ~                       & ~                 & ~               & ~              & Learned & Learned & Learned  & Learned  & Learned  & Learned  & Learned  & Learned & Learned  & Rotary  & ALiBi   & None    & Learned   \\
Activation                  & ~         & ~                       & ~                 & ~               & ~              & GELU    & GELU    & GELU     & GELU     & GELU     & GELU     & GELU     & GELU    & SwiGLU   & GELU    & GELU    & GELU    & GELU      \\
Embedding Norm              & ~         & ~                       & ~                 & ~               & ~              & No      & No      & No       & No       & No       & No       & No       & No      & No       & No      & No      & No      & No        \\
Parameters in billion                 & ~         & ~                       & 1.3               & 6.7             & 1.3            & 1.3     & 1.3     & 1.3      & 1.3      & 1.3      & 1.3      & 1.3      & 13     & 1.3      & 1.3     & 1.3     & 1.3       & 1.3       \\
Tokens
  trained in billion & ~         & ~                       & 300               & 300             & 300            & 112     & 112     & 112      & 250      & 300      & 300      & 330      & 300     & 112      & 112     & 112     & 112     & 112       \\
task                        & metric    & ~                       & ~                 & ~               & ~              & ~       & ~       & ~        & ~        & ~        & ~        & ~        & ~       & ~        & ~       & ~       & ~       & ~         \\
arc\_challenge              & acc       & arc\_challengeacc       & 0.276             & 0.334           & 0.231          & 0.243   & 0.249   & 0.258    & 0.264    & 0.260    & 0.242    & 0.250    & 0.322   & 0.247    & 0.236   & 0.252   & 0.249   & 0.212     \\
arc\_challenge              & acc\_norm & arc\_challengeacc\_norm & 0.295             & 0.375           & 0.259          & 0.274   & 0.261   & 0.275    & 0.277    & 0.286    & 0.277    & 0.290    & 0.342   & 0.268    & 0.270   & 0.276   & 0.260   & 0.243     \\
arc\_easy                   & acc       & arc\_easyacc            & 0.597             & 0.685           & 0.562          & 0.561   & 0.560   & 0.556    & 0.569    & 0.601    & 0.568    & 0.582    & 0.681   & 0.557    & 0.554   & 0.575   & 0.537   & 0.484     \\
arc\_easy                   & acc\_norm & arc\_easyacc\_norm      & 0.555             & 0.633           & 0.502          & 0.503   & 0.478   & 0.506    & 0.518    & 0.528    & 0.516    & 0.515    & 0.600   & 0.502    & 0.476   & 0.491   & 0.461   & 0.434     \\
boolq                       & acc       & boolqacc                & 0.629             & 0.666           & 0.620          & 0.546   & 0.566   & 0.520    & 0.551    & 0.606    & 0.558    & 0.566    & 0.587   & 0.540    & 0.584   & 0.563   & 0.526   & 0.597     \\
copa                        & acc       & copaacc                 & 0.810             & 0.850           & 0.690          & 0.700   & 0.720   & 0.710    & 0.710    & 0.730    & 0.690    & 0.690    & 0.880   & 0.660    & 0.690   & 0.780   & 0.680   & 0.710     \\
hellaswag                   & acc       & hellaswagacc            & 0.429             & 0.504           & 0.387          & 0.422   & 0.404   & 0.374    & 0.385    & 0.405    & 0.378    & 0.380    & 0.542   & 0.379    & 0.410   & 0.422   & 0.395   & 0.340     \\
hellaswag                   & acc\_norm & hellaswagacc\_norm      & 0.545             & 0.664           & 0.489          & 0.551   & 0.515   & 0.464    & 0.486    & 0.521    & 0.477    & 0.476    & 0.716   & 0.475    & 0.524   & 0.549   & 0.495   & 0.424     \\
lambada                     & acc       & lambadaacc              & 0.625             & 0.694           & 0.572          & 0.469   & 0.481   & 0.569    & 0.575    & 0.609    & 0.581    & 0.580    & 0.634   & 0.574    & 0.496   & 0.501   & 0.454   & 0.408     \\
logiqa                      & acc       & logiqaacc               & 0.201             & 0.215           & 0.197          & 0.206   & 0.237   & 0.210    & 0.218    & 0.203    & 0.217    & 0.223    & 0.232   & 0.215    & 0.210   & 0.215   & 0.237   & 0.218     \\
logiqa                      & acc\_norm & logiqaacc\_norm         & 0.269             & 0.292           & 0.273          & 0.267   & 0.270   & 0.275    & 0.286    & 0.269    & 0.281    & 0.280    & 0.275   & 0.272    & 0.254   & 0.272   & 0.293   & 0.283     \\
mathqa                      & acc       & mathqaacc               & 0.244             & 0.251           & 0.241          & 0.233   & 0.222   & 0.249    & 0.248    & 0.263    & 0.246    & 0.245    & 0.238   & 0.245    & 0.234   & 0.237   & 0.215   & 0.223     \\
mathqa                      & acc\_norm & mathqaacc\_norm         & 0.242             & 0.247           & 0.237          & 0.228   & 0.228   & 0.246    & 0.245    & 0.259    & 0.242    & 0.242    & 0.235   & 0.234    & 0.229   & 0.238   & 0.221   & 0.222     \\
mc\_taco                    & f1        & mc\_tacof1              & 0.458             & 0.484           & 0.493          & 0.361   & 0.293   & 0.485    & 0.488    & 0.494    & 0.487    & 0.489    & 0.497   & 0.493    & 0.461   & 0.337   & 0.477   & 0.387     \\
mrpc                        & acc       & mrpcacc                 & 0.578             & 0.684           & 0.684          & 0.684   & 0.588   & 0.684    & 0.684    & 0.684    & 0.679    & 0.679    & 0.677   & 0.684    & 0.684   & 0.684   & 0.679   & 0.302     \\
mrpc                        & f1        & mrpcf1                  & 0.718             & 0.812           & 0.812          & 0.812   & 0.702   & 0.812    & 0.812    & 0.812    & 0.808    & 0.809    & 0.806   & 0.812    & 0.812   & 0.812   & 0.808   & 0.090     \\
multirc                     & acc       & multircacc              & 0.018             & 0.015           & 0.018          & 0.018   & 0.026   & 0.023    & 0.024    & 0.023    & 0.025    & 0.008    & 0.018   & 0.026    & 0.009   & 0.011   & 0.016   & 0.040     \\
openbookqa                  & acc       & openbookqaacc           & 0.224             & 0.290           & 0.216          & 0.220   & 0.200   & 0.190    & 0.196    & 0.222    & 0.194    & 0.208    & 0.294   & 0.214    & 0.212   & 0.224   & 0.210   & 0.170     \\
openbookqa                  & acc\_norm & openbookqaacc\_norm     & 0.336             & 0.386           & 0.336          & 0.336   & 0.328   & 0.316    & 0.314    & 0.334    & 0.302    & 0.312    & 0.412   & 0.320    & 0.344   & 0.340   & 0.332   & 0.276     \\
piqa                        & acc       & piqaacc                 & 0.745             & 0.763           & 0.711          & 0.732   & 0.716   & 0.693    & 0.704    & 0.716    & 0.698    & 0.706    & 0.777   & 0.693    & 0.720   & 0.729   & 0.711   & 0.674     \\
piqa                        & acc\_norm & piqaacc\_norm           & 0.746             & 0.772           & 0.711          & 0.730   & 0.721   & 0.705    & 0.705    & 0.717    & 0.698    & 0.701    & 0.788   & 0.689    & 0.721   & 0.731   & 0.711   & 0.682     \\
prost                       & acc       & prostacc                & 0.270             & 0.288           & 0.238          & 0.243   & 0.237   & 0.249    & 0.229    & 0.204    & 0.219    & 0.226    & 0.281   & 0.244    & 0.287   & 0.280   & 0.240   & 0.253     \\
prost                       & acc\_norm & prostacc\_norm          & 0.260             & 0.295           & 0.308          & 0.293   & 0.303   & 0.268    & 0.271    & 0.268    & 0.292    & 0.305    & 0.283   & 0.276    & 0.296   & 0.332   & 0.300   & 0.313     \\
pubmedqa                    & acc       & pubmedqaacc             & 0.611             & 0.622           & 0.544          & 0.573   & 0.438   & 0.563    & 0.589    & 0.662    & 0.612    & 0.612    & 0.615   & 0.589    & 0.507   & 0.514   & 0.486   & 0.412     \\
qnli                        & acc       & qnliacc                 & 0.512             & 0.529           & 0.499          & 0.476   & 0.507   & 0.505    & 0.506    & 0.505    & 0.499    & 0.499    & 0.517   & 0.498    & 0.493   & 0.481   & 0.493   & 0.493     \\
qqp                         & acc       & qqpacc                  & 0.372             & 0.441           & 0.382          & 0.396   & 0.384   & 0.381    & 0.370    & 0.375    & 0.371    & 0.369    & 0.368   & 0.435    & 0.370   & 0.423   & 0.370   & 0.389     \\
qqp                         & f1        & qqpf1                   & 0.534             & 0.515           & 0.522          & 0.530   & 0.519   & 0.534    & 0.537    & 0.537    & 0.538    & 0.538    & 0.533   & 0.495    & 0.539   & 0.475   & 0.537   & 0.505     \\
race                        & acc       & raceacc                 & 0.356             & 0.386           & 0.341          & 0.330   & 0.323   & 0.334    & 0.329    & 0.344    & 0.321    & 0.323    & 0.374   & 0.337    & 0.317   & 0.344   & 0.332   & 0.326     \\
rte                         & acc       & rteacc                  & 0.585             & 0.552           & 0.603          & 0.502   & 0.534   & 0.563    & 0.549    & 0.578    & 0.563    & 0.549    & 0.524   & 0.527    & 0.545   & 0.524   & 0.527   & 0.505     \\
sciq                        & acc       & sciqacc                 & 0.867             & 0.919           & 0.860          & 0.825   & 0.810   & 0.838    & 0.853    & 0.868    & 0.860    & 0.867    & 0.895   & 0.849    & 0.818   & 0.828   & 0.816   & 0.793     \\
sciq                        & acc\_norm & sciqacc\_norm           & 0.809             & 0.896           & 0.770          & 0.747   & 0.717   & 0.755    & 0.762    & 0.792    & 0.791    & 0.803    & 0.815   & 0.770    & 0.718   & 0.728   & 0.698   & 0.702     \\
sst                         & acc       & sstacc                  & 0.732             & 0.666           & 0.656          & 0.676   & 0.560   & 0.753    & 0.721    & 0.501    & 0.528    & 0.710    & 0.514   & 0.760    & 0.493   & 0.588   & 0.588   & 0.510     \\
triviaqa                    & acc       & triviaqaacc             & 0.115             & 0.195           & 0.052          & 0.027   & 0.025   & 0.056    & 0.065    & 0.058    & 0.047    & 0.049    & 0.133   & 0.050    & 0.031   & 0.039   & 0.028   & 0.021     \\
webqs                       & acc       & webqsacc                & 0.048             & 0.065           & 0.017          & 0.012   & 0.004   & 0.023    & 0.026    & 0.023    & 0.020    & 0.021    & 0.027   & 0.012    & 0.006   & 0.004   & 0.015   & 0.001     \\
wic                         & acc       & wicacc                  & 0.495             & 0.500           & 0.500          & 0.495   & 0.508   & 0.495    & 0.500    & 0.500    & 0.498    & 0.500    & 0.498   & 0.500    & 0.498   & 0.492   & 0.500   & 0.500     \\
winogrande                  & acc       & winograndeacc           & 0.595             & 0.648           & 0.551          & 0.564   & 0.565   & 0.536    & 0.552    & 0.560    & 0.533    & 0.543    & 0.647   & 0.538    & 0.564   & 0.583   & 0.543   & 0.519     \\
wsc                         & acc       & wscacc                  & 0.394             & 0.558           & 0.365          & 0.539   & 0.567   & 0.365    & 0.365    & 0.365    & 0.414    & 0.385    & 0.500   & 0.365    & 0.394   & 0.635   & 0.462   & 0.539     \\
Avg
  acc                   & ~         & ~                       & 45.30\%           & 49.28\%         & 42.94\%        & 42.77\% & 41.72\% & 42.79\%  & 43.12\%  & 43.46\%  & 42.24\%  & 43.08\%  & 47.09\% & 42.95\%  & 41.45\% & 43.70\% & 41.23\% & 38.55\%  
\end{tabular}
\end{tiny}
\end{sidewaystable}

\end{document}